# Compositional Belief Update

**James Delgrande**                                                    JIM@CS.SFU.CA
**Yi Jin**                                                             YIJ@CS.SFU.CA
*School of Computing Science*
*Simon Fraser University, Burnaby, BC, Canada V5A 1S6*

**Francis Jeffry Pelletier**                                           JEFFPELL@SFU.CA
*Departments of Philosophy and Linguistics*
*Simon Fraser University, Burnaby, BC, Canada V5A 1S6*

## Abstract

In this paper we explore a class of belief *update* operators, in which the definition of the operator is compositional with respect to the sentence to be added. The goal is to provide an update operator that is intuitive, in that its definition is based on a recursive decomposition of the update sentence's structure, and that may be reasonably implemented. In addressing update, we first provide a definition phrased in terms of the models of a knowledge base. While this operator satisfies a core group of the benchmark Katsuno-Mendelzon update postulates, not all of the postulates are satisfied. Other Katsuno-Mendelzon postulates can be obtained by suitably restricting the syntactic form of the sentence for update, as we show. In restricting the syntactic form of the sentence for update, we also obtain a hierarchy of update operators with Winslett's *standard semantics* as the most basic interesting approach captured. We subsequently give an algorithm which captures this approach; in the general case the algorithm is exponential, but with some not-unreasonable assumptions we obtain an algorithm that is linear in the size of the knowledge base. Hence the resulting approach has much better complexity characteristics than other operators in some situations. We also explore other compositional belief change operators: *erasure* is developed as a dual operator to update; we show that a *forget* operator is definable in terms of update; and we give a definition of the compositional *revision* operator. We obtain that compositional revision, under the most natural definition, yields the Satoh revision operator.

## 1. Introduction

A knowledge base is typically not a static entity, but rather evolves over time. New information may be added, and old or out-of-date information may be removed. A fundamental issue concerns how such change should be managed. A major body of research addresses this question via the specification of *rationality postulates*, or standards that a change operator should satisfy. These postulates describe belief change at the *knowledge level*, independent of how beliefs are represented and manipulated. There are various rationales for motivating a change in an evolving knowledge base, and these differing rationales have been seen as calling for differences in the background knowledge-level postulates. For example, one may think that some alteration in the world has occurred, with the result that we should *update* the knowledge base's representation of the world in some appropriate way. Or, we may think that our previous sources of information were fallible or incomplete and that we now have better, more accurate information about the world. So, in this case we should *revise* our beliefs. Another motivation might be to *merge* already-existing stores of beliefs, without giving any a priori preference to one or the other of the belief sets, but aiming to achieve a balanced resolution of conflicts. Such a merging might be used to combine the belief states of different agents, so as to come up with a joint course of action based on some sort of "all things considered" assimilation of the knowledge and preferences of the agents that are involved. And we can also imagine a linguistic reform, so that a concept (or rather, the associated word) was no longer to be used. In such a case one might say that the users *forgot* about this concept/word.

These differences in motivation have led to specific differences in the sorts of postulates that are associated with the different motivations. Initially, in the *AGM approach* (Alchourrón, Gärdenfors, & Makinson, 1985; Gärdenfors, 1988), standards for belief *revision* and *contraction* functions were given, wherein it was assumed that a knowledge base is receiving information concerning a static[1] domain, and that it is the increased amount or accuracy of information that is responsible for the changes in the knowledge base. Subsequently, Katsuno and Mendelzon (1992) explored a distinct notion of belief change, with functions for belief *update* and *erasure*, wherein an agent changes its beliefs in response to what it perceives as changes in the environment. The concept of *forget* goes back to George Boole (1854), but was reintroduced in the work of Lin and Reiter (1994) and Lin (2001) as a way to characterize how an agent may bring its knowledge base up-to-date, by forgetting about facts that are no longer relevant and in such a way as to not affect any possible future actions. This approach is syntactic in nature: it deals with the issue of removing facts by removing the ability to *describe* the facts.[2] Finally, the notion of knowledge base merging was introduced as a generalization of the long-standing problem of information sharing between databases, where different databases might contain conflicting information (see Bright, Hurson, & Pakzad, 1992, for a survey). With the work of Revesz (1993), there came an interest in constructing a "merged" knowledge base that best represents the information in a set of other knowledge bases. One use for this was thought to be a way of determining a course of action that best represents the "desires" and "goals" of a divergent set of knowledge bases, thereby forming a group-level, all-things-considered knowledge base. The formal properties of merging have been discussed in previous works (e.g., see Lin & Mendelzon, 1998; Konieczny & Pino Pérez, 1998; Everaere, Konieczny, & Marquis, 2007).

The distinctions between the formal properties of the different types of change were brought out in each of the papers after the initial AGM publications; for instance, Katsuno and Mendelzon (1992) compared update with revision; Konieczny and Pino Pérez (1998) compared merging with revision; Nayak et al. (2006) compared forgetting with update. Some of the postulates suggested by the initial authors of these different conceptions of belief change have been challenged by other writers. And since our own approach towards update conflicts with some of Katsuno and Mendelzon's postulates, we wish to show that this is not, by itself, a reason to reject our theory — every theory has met with objurgation concerning its foundational postulates.

---

1. Note that "static" does not imply "with no mention of time". For example, one could have information in a knowledge base about the state of the world at different points in time, and revise information at these points in time. Thus, belief revision is also applicable to the situation where an agent investigates a past event and tries to reason about what was the real state of the world when this event took place. Further considerations on how revision and update are interrelated are in the work of Lang (2006).

2. Nayak, Chen, and Lin (2006) described this difference thus: "While belief erasure purports to answer the question 'What should I believe if I can no longer support the belief that the cook killed Cock Robin?', forgetting purports to answer the question 'What should I believe if *Killing* was a concept not afforded in my language?'."

Although our focus in this paper is with update — and hence with the postulates given by Katsuno and Mendelzon (1992) and the objections related to these postulates — we believe that considerations similar to the ones we bring forward in this arena would hold with respect to the other sorts of belief change postulates. That is, we think that the rationale we have for imposing a compositionality constraint on belief update should be brought to bear on the cases of belief revision, belief merging, and forgetting.

The knowledge level specifications of these types of belief change allow for different ways to implement any of them. Various researchers have proposed specific change operators for belief revision (Borgida, 1985; Dalal, 1988; Satoh, 1988), belief update (Forbus, 1989; Weber, 1986; Winslett, 1988), belief merging (Subrahmanian, 1994; Konieczny, 2000; Everaere, Konieczny, & Marquis, 2005), and forgetting (Lang, Liberatore, & Marquis, 2003; Nayak et al., 2006). These approaches are formulated in terms of the distance between models of the knowledge base and models of a sentence for revision or update. In general there has been less work dealing with systems that may be readily implementable (but see, e.g., Williams, 1996; Delgrande & Schaub, 2003).

In this paper we develop a specific update operator where the operator is intended to be *compositional*, in that an update $\psi \diamond \mu$ can be expressed recursively in terms of the syntactic structure of $\mu$. Thus, if a knowledge base is to be updated by a disjunction $\mu = a \vee b$, the idea is that this update will be a function of the update by $a$ in a certain combination with the update by $b$. The update of the knowledge base by a conjunction $\mu = a \wedge b$ will also be a function (a different one) of the update by $a$ in combination with the update by $b$. The goal is to arrive at an operator whose results are intuitive, in that its definition is based on a recursive decomposition of a formula; hence the (generally abstract) notion of update will be anchored in part in a more familiar computational setting. Second, the hope is that these operators will be efficiently implementable, at least in some cases, by exploiting restrictions to the syntactic form of the formula. The focus here is on the form of the formula for update; presumably the approach described may be combined with one in which the knowledge base is itself divided into relevant and irrelevant parts for an update (Parikh, 1999).

These goals are generally realised. First, the operators have reasonable properties: many of the Katsuno and Mendelzon benchmark properties are satisfied, including those deemed essential by Herzig and Rifi (1999). While we don't obtain full irrelevance of syntax, we do obtain weaker results in this regard; as well we show how irrelevance of syntax can be obtained by restricting the syntactic form of the sentence for update. The approach is also related to other approaches in the literature, and hence serves to establish some links between approaches. In fact, the family of compositional update operators obtained by imposing various syntactic restrictions can be regarded as constituting a family of operators of which Winslett's *standard semantics* makes up the most basic nontrivial approach. As well, the general approach to update presented here can capture the *forget* operator (Lin & Reiter, 1994; Lang et al., 2003; Nayak et al., 2006), and so in a certain sense can be regarded as generalizing *forget*. We also define a revision operator using the obvious definition for such an operator; it proves to be the case that this operator corresponds with the revision operator in the work of Satoh (1988).

The approach leads to a straightforward algorithm for implementing these operators. This algorithm is efficient, compared to the model-based definition of this and other distance-based operators. For a knowledge base in disjunctive normal form, the size of the knowledge base contributes only a linear factor to the overall complexity. As well, further efficiency is obtained when the size of the input sentence is bounded by a constant.

The next section reviews belief revision, update, forgetting, and merging, and describes two specific approaches to update. The section following describes our approach, after which, in the next section, we give a discussion and analysis. The last section contains concluding remarks; proofs of theorems are given in an Appendix.

## 2. Background

As described, our goal is to introduce a compositional method of carrying out belief change. But since part of our overall goal also is to examine the place of a compositional belief change operation in all the various arenas where this can take place, we start by outlining some of the details for each of these different conceptions that motivate belief change, along with some motivational considerations and some areas where the different types of belief change part ways. These operators were introduced *implicitly*, by means of a set of postulates that any legitimate such operator was required to obey. However, in all these areas there has been some dispute concerning the correctness of the various postulates, and we mention some of these as we proceed, since our own approach in the case of update does not obey all the standard postulates for update. We start with the historically earlier case of revision before moving to our central concern of update. These are followed by short expositions concerning forgetting and merging.

### 2.1 Formal Preliminaries

We consider a propositional language $L$, over a finite set of atoms, that is, propositional letters, $\mathbf{P} = \{\top, a, b, c, \dots\}$, and truth-functional connectives $\neg$, $\wedge$, and $\vee$. Where convenient, $\supset$ and $\equiv$ are also used, and are considered as being introduced by definition. We use $\leftrightarrow$ for *logical equivalence*; that is, $\alpha \leftrightarrow \beta$ is an abbreviation for $\vdash (\alpha \equiv \beta)$. *Lits* is the set of literals $\mathbf{P} \cup \{\neg l \mid l \in \mathbf{P}\}$. In particular, $\neg\top$ is also denoted as $\bot$. A set of literals $\Gamma$ is consistent just if $\bot \notin \Gamma$ and for no atom $p \in \mathbf{P}$ do we have $p, \neg p \in \Gamma$. For a literal $l$, we use $\bar{l}$ to denote $\neg l$ if $l \in \mathbf{P}$, or $l' \in \mathbf{P}$ if $\neg l' = l$. Similarly, for a set of literals $\Gamma$, we use $\overline{\Gamma}$ to denote the set $\{\bar{l} \mid l \in \Gamma\}$. The expression $atom(\mu)$ denotes the set of atoms in formula $\mu$. An *interpretation* $\omega$ of $L$ is a maximal consistent set of literals, i.e., $\top \in \omega$ and for every other $p \in \mathbf{P}$ precisely one of $p \in \omega$, $\neg p \in \omega$ holds. A *model* of a sentence $\mu$ is an interpretation that makes $\mu$ true, according to the usual definition of truth. $Mod(\mu)$ denotes the set of models of sentence $\mu$. We also make use of the notation $ModL(\mu)$ to denote the set of models of sentence $\mu$ over the language of $\mu$ (that is to say, over the language $atom(\mu)$.) For interpretation $\omega$ we write $\omega \models \mu$ to mean $\mu$ is true in $\omega$. For interpretation $\omega$ and set of literals $\Gamma$, we define $\omega \downarrow \Gamma = \omega \setminus (\Gamma \cup \overline{\Gamma})$. That is, $\omega \downarrow \Gamma$ is the set of literals in $\omega$ but containing neither $l$ nor $\bar{l}$ for each $l \in \Gamma$. For example, if $\omega = \{a, \neg b, \neg c\}$ then $\omega \downarrow \{b, \neg c\} = \{a\}$.

We denote the negation-normal form (in which negation applies to atoms only) of a sentence $\mu$ by $nnf(\mu)$. Similarly, we denote the conjunctive normal form and the disjunctive normal form of $\mu$ by $cnf(\mu)$ and $dnf(\mu)$ respectively.[3] For a set of sentences $\Gamma$ (which will always be finite), we use $\bigvee \Gamma$ to denote the disjunction and $\bigwedge \Gamma$ the conjunction of the sentences in $\Gamma$. Proofs will often be based on the structure of a formula, specifically on the *depth* of a formula; for formula $\mu$, the *depth* of $\mu$, $depth(\mu)$ is the maximum nesting of connectives in $\mu$. Hence $depth(\neg a \vee (b \wedge \neg c)) = 3$.

Later we make use of the notion of the *prime implicants* of a sentence. A consistent set of literals $\Gamma$ is a *prime implicant* of $\mu$ iff: $\Gamma \vdash \mu$ and for any $\Gamma' \subset \Gamma$ we have $\Gamma' \not\vdash \mu$.[4] In the limiting case where $\vdash \mu$, we take the (sole) prime implicant of $\mu$ to be $\{\top\}$.

## 2.2 Belief Revision and Contraction

In the seminal approach of AGM (Alchourrón et al., 1985), postulates are proposed to constrain belief revision. In this approach, a knowledge base $K$ is assumed to be a *belief set*, a set of sentences closed under logical consequence. The revision of a belief set by a formula, $K * \phi$, is a new belief set in which the formula $\phi$ is believed. The interesting case is that in which $\neg \phi$ is initially believed, and so to attain a consistent belief set (assuming that $\phi$ is satisfiable), some beliefs have to be dropped. Exactly which beliefs must be dropped is not stipulated in the AGM approach; however, constraints in the form of postulates that govern what are seen as legitimate revision operators are given. In contrast, in their development of belief update Katsuno and Mendelzon (1992) represented the knowledge base by a *formula* in some language $L$. Hence, in this paper we also express things in terms of postulates phrased in terms of formulas, rather than belief sets.

The following **R**-postulates comprise Katsuno and Mendelzon's reformulation of the AGM revision postulates, where $*$ is a function from $L \times L$ to $L$.

**(R1)** $\psi * \mu \vdash \mu$.

**(R2)** If $\psi \wedge \mu$ is satisfiable, then $\psi * \mu \leftrightarrow \psi \wedge \mu$.

**(R3)** If $\mu$ is satisfiable then $\psi * \mu$ is also satisfiable.

**(R4)** If $\psi_1 \leftrightarrow \psi_2$ and $\mu_1 \leftrightarrow \mu_2$ then $\psi_1 * \mu_1 \leftrightarrow \psi_2 * \mu_2$.

**(R5)** $(\psi * \mu) \wedge \phi \vdash \psi * (\mu \wedge \phi)$.

**(R6)** If $(\psi * \mu) \wedge \phi$ is satisfiable then $\psi * (\mu \wedge \phi) \vdash (\psi * \mu) \wedge \phi$.

A dual operation, called *contraction* is also defined, in which a formula is deleted from the knowledge base. This operation can be seen as governed by the **C**-postulates, again using a Katsuno and Mendelzon formulation in terms of a function from $L \times L$ to $L$.

---

3. Of course for formula $\mu$, there are many different but logically equivalent ways to express $cnf(\mu)$ and $dnf(\mu)$. We assume a fixed procedure for converting to cnf (or dnf), by converting to negation normal form, and then distributing disjunctions over conjunctions (or vice versa for dnf), hence justifying the use of the term *the* conjunctive (disjunctive) normal form of a formula, rather than *a* (disjunctive) normal form.

4. The notion of prime implicant should not be confused with the dual notion of a *prime implicate*. A prime implicate of a formula $\mu$ is a clause, or disjunction of literals, $\rho$, such that $\mu \vdash \rho$ but for any proper subclause $\rho'$ of $\rho$, we have $\mu \not\vdash \rho'$.

**(C1)** $\psi \vdash \psi - \mu$.

**(C2)** If $\psi \nvdash \mu$ then $\psi - \mu \leftrightarrow \psi$.

**(C3)** If $\nvdash \mu$ then $\psi - \mu \nvdash \mu$.

**(C4)** If $\psi_1 \leftrightarrow \psi_2$ and $\mu_1 \leftrightarrow \mu_2$ then $\psi_1 - \mu_1 \leftrightarrow \psi_2 - \mu_2$.

**(C5)** $(\psi - \mu) \wedge \mu \vdash \psi$.

Revision and contraction are related in the AGM approach by what have come to be known as the *Levi* and *Harper* identities. They may be expressed as follows (using formulas rather than belief sets):

$$\psi * \mu \quad \leftrightarrow \quad (\psi - \neg\mu) \wedge \mu \tag{1}$$

$$\psi - \mu \quad \leftrightarrow \quad \psi \vee (\psi * \neg\mu). \tag{2}$$

The first case asserts that revising $\psi$ by $\mu$ corresponds to the contraction of $\psi$ by $\neg\mu$ conjoined with $\mu$. The second asserts that contracting $\mu$ from $\psi$ corresponds to the disjunction of $\psi$ with the result of $\psi$ updated by $\neg\mu$.

Although this makes a nice picture, there have been various objections to some of the presuppositions of the AGM model (e.g., the representation of belief states by *theories*, that is, by infinite sets of formulas) and to some of the postulates that are said to govern the operations of revision and contraction (especially **(C5)**, the postulate of "recovery"). Issues involved with **(C5)** have been discussed by Fuhrmann, 1991; Tennant, 1997; Hansson & Rott, 1998; Rott & Pagnucco, 1999, and others.

### 2.3 Belief Update and Erasure

The account of revision and contraction described in the preceding subsection is usually seen as applying most straightforwardly to the case where one has a store of information about "an unchanging, static world" but where new information about that world is received by the agent, thereby forcing a change in the representation of this "unchanging, static world." But a different picture was put forward by Katsuno and Mendelzon (1992), where there was a "changing, dynamic world". In such a conception, the new information that is gathered by the agent reflects the idea that the world is different than it was when the knowledge base was previously constructed. The sorts of changes to the knowledge base that are required by this type of new information are seen as different from the sorts envisaged when it is thought that changes to the knowledge base are only going to make its contents successively more accurate. Although this simplistic distinction is not all there is to the differences between the two pictures (as we mentioned in Footnote 1), it has led to a large body of work that does point to a different conception. Distinct operations that change knowledge bases have been proposed: *update*, which makes changes to the knowledge base given information concerning a change in the state of the world, and *erasure*, for removing out-of-date information.

A formula is said to be *complete* just if it implies the truth or falsity of every other formula. In the approach of (Katsuno & Mendelzon, 1992), *update* is a function $\diamond$ from $L \times L$ to $L$ satisfying the following **U**-postulates.

**(U1)** $\psi \diamond \mu \vdash \mu$.

**(U2)** If $\psi \vdash \mu$ then $(\psi \diamond \mu) \leftrightarrow \psi$.

**(U3)** If $\mu$ and $\psi$ are satisfiable then so is $\psi \diamond \mu$.

**(U4)** If $\psi_1 \leftrightarrow \psi_2$ and $\mu_1 \leftrightarrow \mu_2$ then $(\psi_1 \diamond \mu_1) \leftrightarrow (\psi_2 \diamond \mu_2)$.

**(U5)** $(\psi \diamond \mu) \wedge \phi \vdash \psi \diamond (\mu \wedge \phi)$.

**(U6)** If $\psi \diamond \mu_1 \vdash \mu_2$ and $\psi \diamond \mu_2 \vdash \mu_1$ then $(\psi \diamond \mu_1) \leftrightarrow (\psi \diamond \mu_2)$.

**(U7)** If $\psi$ is complete then $(\psi \diamond \mu_1) \wedge (\psi \diamond \mu_2) \vdash \psi \diamond (\mu_1 \vee \mu_2)$.

**(U8)** $(\psi_1 \vee \psi_2) \diamond \mu \leftrightarrow (\psi_1 \diamond \mu) \vee (\psi_2 \diamond \mu)$

These postulates are not, however, uncontentious. Herzig and Rifi (1999) discussed the plausibility of the postulates given; they assert that **U2**, **U5**, and **U6** are undesirable, while **U7** is unimportant. This leaves (according to the authors) **U1**, **U3**, **U4**, and **U8** as being desirable.

*Erasure* is also defined, in a manner analogous to the way we described how contraction was related to belief revision. In both cases, some specified formula is not believed in the result. The erasure of $\mu$ from $\psi$ is denoted $\psi \blacksquare \mu$, and the formula $\mu$ is not believed in the resulting state. As with all our other operations, there is a set of postulates characterizing erasure (given in Katsuno & Mendelzon, 1992). Update and erasure are also interdefinable by means of identities, analogous to the *Levi* and *Harper* identities, which related revision and contraction:

$$\psi \diamond \mu \quad \leftrightarrow \quad (\psi \blacksquare \neg \mu) \wedge \mu \tag{3}$$

$$\psi \blacksquare \mu \quad \leftrightarrow \quad \psi \vee (\psi \diamond \neg \mu). \tag{4}$$

The first case asserts that update by $\mu$ corresponds to erasing $\neg \mu$ along with the conjunction with $\mu$. The second asserts that erasing $\mu$ from $\psi$ corresponds to disjoining $\psi$ with the result of $\psi$ updated by $\neg \mu$.

There have been various specific update (and revision) operators proposed based on the distance between interpretations. We focus on two update operators, both due to Winslett. The first, the *Possible Models Approach (PMA)* of (Winslett, 1988) is a well-known example of an update operator satisfying the Katsuno and Mendelzon update postulates. The second, the *standard semantics* of (Winslett, 1990) is a weak (in fact, arguably *the* weakest reasonable) approach to update. We denote these operators by $\diamond_{pma}$ and $\diamond_{ss}$ respectively.

For $\psi \diamond_{pma} \mu$, we have that, for each interpretation $w$ of $\psi$, $\diamond_{pma}$ selects from the interpretations of $\mu$ those that are "closest" to $w$. The update is determined by the set of these closest interpretations. The notion of "closeness" between two interpretations $w_1$ and $w_2$ is the Hamming distance, given as follows:

**Definition 1** *$diff(w_1, w_2) =$ The set of all propositional letters on which $w_1$ and $w_2$ differ.*

Interpretation $w_1$ is not less close to $w$ than $w_2$, $w_1 \leq_w w_2$, just if $diff(w, w_1) \subseteq diff(w, w_2)$. It follows that $\leq_w$ is a partial order on interpretations. The $\leq_w$-minimal set with respect to $\mu$ is designated $Min(Mod(\mu), w)$. ¿From this we can specify the PMA update operator:

$$Mod(\psi \diamond_{pma} \mu) = \bigcup_{w \in Mod(\psi)} Min(Mod(\mu), w).$$

The update operator $\psi \diamond_{ss} \mu$ is defined so that for each model of $\psi$, those models of $\mu$ that retain the truth values of atoms not in $\mu$ are chosen. That is:

$$Mod(\psi \diamond_{ss} \mu) = \bigcup_{w_1 \in Mod(\psi)} \{w_2 \in Mod(\mu) \mid diff(w_1, w_2) \subseteq atom(\mu)\}$$

The operator $\psi \diamond_{ss} \mu$ is the weakest "reasonable" update operator in the following sense (Winslett, 1990): First, for an update $\psi \diamond_{ss} \mu$, $\mu$ is true in every model of $\psi \diamond_{ss} \mu$. Second, every model of $\psi$ over the language excluding atoms in $\mu$ is a model of $\psi \diamond_{ss} \mu$ (again over this restricted language). Moreover, $\psi \diamond_{ss} \mu$ consists of the maximal set of interpretations that satisfies the preceding two properties. Hence in the update of $\psi$ by $\mu$, the truth values of atoms in $\psi$ but not in $\mu$ are unaffected by the update.

**Example 1 (Katsuno & Mendelzon, 1992)** *Let $L = \{b, m\}$ be the language of discourse. Let $\psi = (b \wedge \neg m) \vee (\neg b \wedge m)$, and $\mu = b$. The interpretations of $\psi$ are $w_1 = (\neg b, m)$, $w_2 = (b, \neg m)$; and the interpretations of $\mu$ are: $w_1' = (b, m)$, $w_2' = (b, \neg m)$. Thus $diff(w_1, w_1') = \{b\}$ and $diff(w_1, w_2') = \{b, m\}$, hence $w_1' \leq_{w_1} w_2'$ and $w_2' \not\leq_{w_1} w_1'$, so $Min(Mod(\mu), w_1) = \{w_1'\}$. Similarly, $Min(Mod(\mu), w_2) = \{w_2'\}$. Hence, $(\psi \diamond_{pma} \mu) \leftrightarrow b$. The same result obtains for $\diamond_{ss}$.*

For concreteness, take $b$ to mean "the book is on the floor", and $m$ to mean "the magazine is on the floor". So $\psi$ means that either the book or the magazine is on the floor, but not both. A robot is ordered to put the book on the floor. Intuitively, at the end of this action the book will be on the floor, and the location of the magazine will be unknown. Both operators give this result.

**Example 2** *Let $\psi = (\neg b \wedge \neg m)$ and $\mu = (b \vee m)$. Then $(\psi \diamond_{pma} \mu) \leftrightarrow (b \equiv \neg m)$, whereas $(\psi \diamond_{ss} \mu) \leftrightarrow (b \vee m)$.*

Here, neither the book nor the magazine is on the floor. The robot is ordered to put at least one of them on the floor. According to the $\diamond_{pma}$ operator, exactly one will be on the floor after this action, while according to the $\diamond_{ss}$ operator, at least one will be on the floor.

## 2.4 Forget

While our focus is on a specific approach to update and erasure, we also relate our approach to that of the *forget* operator. The notion of forgetting goes back to George Boole (1854), though it has received more recent attention in Artificial Intelligence by, e.g., Lin & Reiter, 1994; Lin, 2001; Lang et al., 2003; Nayak et al., 2006. In a propositional context, to forget an atom, or set of atoms, is to remove all information concerning the atom or set of atoms.

It has been suggested (in Nayak et al., 2006) that forgetting corresponds to the removal of literals or atoms from the language of discourse in the case of propositional forgetting (i.e., 0-place predicate forgetting). In the more general case, it is seen as removing a predicate or relation from the language, and hence removing any further consequences that might have been due to this predicate's presence.

Let $\psi[p/q]$ denote the formula $\psi$ where all occurrences of atom $p$ are replaced by $q$. Then the usual definition for forgetting (again, going back to Boole) atom $p$ in $\psi$ is given by $\psi[p/\top] \vee \psi[p/\bot]$. In order to forget a set of atoms $\Gamma$, one takes the disjunction of the substitution of all $2^{|\Gamma|}$ combinations of $\top$, $\bot$ for elements of $\Gamma$.

We have the following definitions. For single atoms we basically follow Nayak et al. (2006); for sets of atoms we use the definition from (Lin & Reiter, 1994). To begin, the *p-dual* of an interpretation $\omega$ is the interpretation like $\omega$ but where the truth value assigned to $p$ is changed to its negation. A set of interpretations is closed under $p$-duals just if, for any interpretation $\omega$ in the set, the $p$-dual of $\omega$ is also in the set.

**Definition 2** *Given a set of interpretations $\Omega$ and atom $p$, the operator $\biguplus(\Omega, p)$ yields the least set of interpretations containing $\Omega$ and closed under p-duals.*

Given this, we can define forget for an atom and set of atoms, where the latter is defined recursively in terms of the former:

**Definition 3** *Basis Case:* *Let $\psi$ be a formula and $p$ an atom. Then* forget *of $p$ with respect to $\psi$ is given by:*

$$
\begin{aligned}
Mod(\psi \odot p) &= \biguplus(Mod(\psi), p) \\
&= Mod(\psi[p/\top] \vee \psi[p/\bot]).
\end{aligned}
$$

*Inductive Case:* *Let $\psi$ be a formula and $\Gamma = \{p_1, \ldots, p_n\}$ a set of atoms. Then* forget *of $\Gamma$ with respect to $\psi$ is given by:*

$$
\psi \odot \Gamma = (\psi \odot (\Gamma \setminus \{p_n\})) \odot p_n.
$$

For example, $a \wedge (b \vee c) \odot a \leftrightarrow \neg a \wedge (b \vee c) \odot a \leftrightarrow b \vee c$. (Given a knowledge base that has stored that Alberta is in Canada and also that either Vancouver is in British Columbia or Charlottetown is in Ontario, forgetting that Alberta is in Canada would yield that either Vancouver is in British Columbia or Charlottetown is in Ontario. This would be the same result if the initial knowledge base had that Alberta was not in Canada, but that either Vancouver is in British Columbia or Charlottetown is in Ontario.) For another example, $(a \vee b) \odot a \leftrightarrow \top$. This last example illustrates that forget is distinct from erasure, since a property of erasure is that if $\psi$ does not imply $\mu$ then $(\psi \blacksquare \mu) \leftrightarrow \psi$ (Katsuno & Mendelzon, 1992).

## 2.5 Belief Merging

Merging differs formally from the preceding three pictures of how knowledge bases are changed. The preceding operators had a knowledge base and a sentence that may need to occasion a change in the knowledge base. If one rephrases this in terms of agents, these other types of change postulate an agent, with a store of beliefs, who is now faced with a new belief that needs to be accommodated. In the case of merging, however, we start with many belief sets that need all to be dealt with in some way that yields "the best, overall" single belief state. In terms of agents, again, we have here a number of agents, each with a belief set, and we are trying to construct that belief set which best represents the total beliefs of the community of agents. So, rather than being a function that maps a belief set and a sentence onto a belief set, it is instead a function that maps a number of belief sets into a single one. Following our earlier practice of representing belief sets by a single formula (in the manner of Katsuno & Mendelzon, 1992), we can see that the earlier rationales for belief change envision it as a function $L \times L \to L$, whereas merging envisions a function $L \times L \times \ldots \times L \to L$. Note that the general case allows for some of the knowledge bases on this list to be identical to one another, thus the list is actually a multi-set (bag).

The goal in merging, then, is to construct, from a finite list of knowledge bases $E$, some appropriate, single "merged" knowledge base. Despite this formal difference from the earlier three types of belief change, we nevertheless include a discussion here because of the conceptual similarities that hold between merging and any of the other versions of belief change. Indeed, it seems plausible to suggest that merging might be definable in terms of the others, or maybe that it is some sort of generalization of the others. In these cases, our considerations about compositionality of belief change operators may be relevant.

**Definition 4** *A knowledge set is a multi-set (bag) of knowledge bases.*

**Definition 5** *If $E$ is a knowledge set, then $\bigwedge E$ is the conjunction of the formulas representing all the knowledge bases that are in $E$.*

Konieczny and Pino Pérez (1998, 2002) proposed the following **M**-principles to govern all merging operators. A *merge* function $\triangle$ is a function from a knowledge set E to a knowledge base $\triangle(E)$ satisfying the following postulates, where $\sqcup$ is multiset union.[5]

**(M1)** $\triangle(E)$ is consistent

**(M2)** If $\bigwedge E$ is consistent then $\triangle(E) = \bigwedge E$.

**(M3)** If $E_1$ and $E_2$ are knowledge sets such that $E_1 \leftrightarrow E_2$, then $\triangle(E_1) \leftrightarrow \triangle(E_2)$

**(M4)** If $K_1$ and $K_2$ are knowledge bases that are not mutually consistent, then $\triangle(K_1 \sqcup K_2) \not\vdash K_1$

**(M5)** $\triangle(E_1) \wedge \triangle(E_2) \vdash \triangle(E_1 \sqcup E_2)$

**(M6)** If $\triangle(E_1) \wedge \triangle(E_2)$ is consistent, then $\triangle(E_1 \sqcup E_2) \vdash \triangle(E_1) \wedge \triangle(E_2)$

Some of these merging postulates have been contested: For example, Meyer (2000) argued that **M4** and **M6** should be rejected. (He argues this on the grounds that there are many plausible merging operations that do not obey these postulates).

---

5. For simplicity, we list the postulates of (Konieczny & Pino Pérez, 1998), which do not include integrity constraints.

A natural method for determining whether a formula $\phi$ should be in the merged knowledge base is to determine whether it appears in the majority of the members of the knowledge set that is being merged ("the merged knowledge base should allow the opinion of the majority to prevail"). Liberatore and Schaerf (1998) introduced the method of *arbitration*, whereby the goal is to adopt as many different opinions as possible from the members of the knowledge set ("try to take as many differing opinions as possible into account"). Konieczny and Pino Pérez (1998) proved that there is no arbitration operator (at least, not of the sort that they characterize) that obeys **M1** − **M6**.[6] The interplay between various merging operations and the ability of an agent to hide, lie, or otherwise camouflage its preferences from other agents as they try to construct a merged knowledge base has been surveyed in Everaere et al. (2007).

## 3. The Approach

This section discusses our approach. Following intuitions and motivation of the formal approach, we introduce compositional update and, subsequently, erasure. We also consider the notion of compositional belief revision, but conclude that, at least with respect to our specific approach, there is no separate, distinct, notion of compositional revision. Analysis of properties of these operators is covered in the next section.

### 3.1 Intuitions

Our goal is to define update operators in a compositional fashion so that, for updating by formula $\mu$, update is defined in terms of the syntactic components of $\mu$. The general idea behind update is that for $\psi \diamond \mu$, each model of $\psi$ is replaced by the "closest" model(s) in $\mu$ (Katsuno & Mendelzon, 1992). In our approach, the notion of "close" for each model of $\psi$ is determined in part by the syntactic structure of $\mu$. That is, $\mu$ is recursively decomposed; the resulting (base case) literals are used to determine models of the update by sets of literals; and the results are combined depending on the connective(s) in $\mu$.

Consider how this may be carried out. We are given a knowledge base $\psi$ and a sentence $\mu$, and we wish to determine a new knowledge base where $\mu$ is believed. For a base case, $\mu = l$ is a literal, and we wish to update the knowledge base $\psi$ by literal $l$. If $\psi$ implies $l$ then we need do nothing. If $\psi$ does not imply $l$, then we wish to arrive at a knowledge base in which $l$ is believed. That is, we want to change the knowledge base only enough so that it entails $l$. Clearly, we can do this by replacing each model $\omega$ of $\psi$ by the interpretation $\omega' = (\omega \downarrow \{l\}) \cup \{l\}$.[7] Thus, we would have that every resulting interpretation entails $l$.

Consider next updating a knowledge base $\psi$ by a conjunction of literals $\mu = l_1 \wedge l_2$. A knowledge base in which $l_1 \wedge l_2$ is believed will, obviously, be one in which every model of the knowledge base entails both $l_1$ and $l_2$. We carry this out by replacing each interpretation $\omega \in Mod(\psi)$ with an interpretation $\omega' = (\omega \downarrow \{l_1, l_2\}) \cup \{l_1, l_2\}$. There is a limiting case that needs to be taken care of, where $l_1$ is $\overline{l_2}$. In this situation, there is no interpretation in which $l_1$, $l_2$ are true, and in this case $\omega'$ does not exist, reflecting an attempt to update by an inconsistent formula.

---

6. This forms a part of the rationale for Meyer (2000, 2001) to deny **M4** and **M6**.

7. To be clear, if $\omega \models l$ then $\omega' = \omega$; and if $\omega \not\models l$ then $\omega'$ is like $\omega$ but with $l$ replacing its complement.

To update a knowledge base $\psi$ by a disjunction of literals $\mu = l_1 \vee l_2$, we want to modify models of $\psi$ so that at least one of $l_1$ or $l_2$ is true. Consider $\omega \in Mod(\psi)$ such that $\omega \not\models l_1 \vee l_2$. Then $\omega_1 = (\omega \downarrow \{l_1\}) \cup \{l_1\}$ is an interpretation that involves the least change to $\omega$ in which $l_1$ is true, while $\omega_2 = (\omega \downarrow \{l_2\}) \cup \{l_2\}$ does the same for $l_2$. Arguably then $\omega$ should be replaced by $\omega_1$ and $\omega_2$.

Last, we generalize the above considerations to deal with arbitrary formulas. So to update by a disjunction of formulas, we recursively determine the update given by the individual disjuncts and return the union of the resulting sets of interpretations.

## 3.2 A Compositional Update Operator

Based on the preceding intuitions, we define an update operator $\diamond_c$. We begin with some preliminary definitions. In the following, $UL$ is a function from an interpretation $\omega$ and finite set of formulas $\Gamma$ to a set of interpretations. Informally, $\omega$ is a model of the knowledge base and $\Gamma$ is a set of formulas resulting from the partial decomposition of a formula for update. The value of $UL$ is the set of interpretations closest to $\omega$, according to $\Gamma$. To ease notation, in the case of a single formula we sometimes write $UL(\omega, \mu)$ for $UL(\omega, \{\mu\})$.

**Definition 6** *For interpretation $\omega$ and finite $\Gamma \subseteq L$, define $UL(\omega, \Gamma)$ as follows:*

1. *If $\Gamma \subseteq Lits$ then*

$$UL(\omega, \Gamma) = \begin{cases} \{(\omega \downarrow \Gamma) \cup \Gamma\} & \text{if } \Gamma \not\vdash \bot \\ \emptyset & \text{otherwise} \end{cases}$$

2. *If $\Gamma = \{\alpha \wedge \beta\} \cup \Gamma'$ then $UL(\omega, \Gamma) = UL(\omega, \{\alpha, \beta\} \cup \Gamma')$*

3. *If $\Gamma = \{\alpha \vee \beta\} \cup \Gamma'$ then $UL(\omega, \Gamma) = UL(\omega, \{\alpha\} \cup \Gamma') \cup UL(\omega, \{\beta\} \cup \Gamma')$*

4. *If $\Gamma = \{\neg(\alpha \vee \beta)\} \cup \Gamma'$ then $UL(\omega, \Gamma) = UL(\omega, \{\neg\alpha, \neg\beta\} \cup \Gamma')$*

5. *If $\Gamma = \{\neg(\alpha \wedge \beta)\} \cup \Gamma'$ then $UL(\omega, \Gamma) = UL(\omega, \{\neg\alpha\} \cup \Gamma') \cup UL(\omega, \{\neg\beta\} \cup \Gamma')$*

6. *If $\Gamma = \{\neg\neg\alpha\} \cup \Gamma'$ then $UL(\omega, \Gamma) = UL(\omega, \{\alpha\} \cup \Gamma')$*

It is worth noting that the recursion steps of the above definition resemble closely the procedure which we use to convert a formula to its disjunctive normal form. Before defining update in terms of this operator, we first investigate some of its properties. Foremost, we need to show that $UL$ is well-defined. That is, in specifying $UL(\omega, \Gamma)$, the definition is phrased in terms of some member of $\Gamma$; it needs to be shown that the order in which elements are "selected" in the recursion does not affect the result.

**Theorem 1** *$UL$ is well-defined.*

The next two results reflect the influence of the structure of a formula on the recursive decomposition in the definition of $UL$.

**Theorem 2** *$UL(\omega, \Gamma) = UL(\omega, nnf(\bigwedge \Gamma))$.*

**Theorem 3** *$UL(\omega, \Gamma) = UL(\omega, dnf(\bigwedge \Gamma))$.*

Note that a similar result does not extend to conjunctive normal form. A counterexample is given by the following:

$$
\begin{aligned}
UL(\omega, \{a \vee (b \wedge c)\}) &= UL(\omega, \{a\}) \cup UL(\omega, \{b, c\}) \\
&\neq UL(\omega, \{a\}) \cup UL(\omega, \{a, c\}) \cup UL(\omega, \{b, a\}) \cup UL(\omega, \{b, c\}) \\
&= UL(\omega, \{a, a \vee c\}) \cup UL(\omega, \{b, a \vee c\}) \\
&= UL(\omega, \{(a \vee b), (a \vee c)\}) \\
&= UL(\omega, \{(a \vee b) \wedge (a \vee c)\}).
\end{aligned}
$$

We consider next a couple of fundamental properties of $UL$:

**Theorem 4** *For every $\mu \in \Gamma$ and $w' \in UL(\omega, \Gamma)$ we have $w' \models \mu$.*

**Theorem 5** $UL(\omega, \Gamma) = \emptyset$ *iff* $\Gamma \vdash \bot$.

We next define our update operator directly in terms of $UL$.

**Definition 7**
$Mod(\psi \diamond_c \mu) = \{\omega' \mid \omega' \in UL(\omega, \{\mu\}), \omega \in Mod(\psi)\}$.

Recall Example 1 in which $\mu = b$ and $\psi = (b \wedge \neg m) \vee (\neg b \wedge m)$. We have that $Mod(\psi \diamond_c \mu) = \{\omega' \mid \omega' \in UL(\omega, \{\mu\}), \omega \in Mod(\psi)\} = \{(\omega \downarrow \{b\}) \cup \{b\} \mid \omega \in Mod(\psi)\}$. Thus, $Mod(\psi \diamond_c \mu) = \{\{b, \neg m\}, \{b, m\}\}$, and so $(\psi \diamond_c \mu) \leftrightarrow b$. This is the same result as we obtain with both Winslett's approaches.[8]

For Example 2, where $\psi = \neg b \wedge \neg m$ and $\mu = (b \vee m)$, we obtain $Mod(\psi \diamond_c \mu) = \{\{b, \neg m\}, \{\neg b, m\}\}$. In this case, our update operator behaves the same as $\diamond_{pma}$, but differently from $\diamond_{ss}$.

We can similarly define an erasure operator via $UL$. To erase $\mu$ from $\psi$, and in analogy to the Harper Identity, one can update by $\neg \mu$ and add the result to $\psi$. Thus:

**Definition 8**
$Mod(\psi \bullet_c \mu) = Mod(\psi) \cup \{\omega' \mid \omega' \in UL(\omega, \{\neg \mu\}), \omega \in Mod(\psi)\}$.

We get the results:

**Theorem 6**

$$
\begin{aligned}
\psi \diamond_c \mu &\leftrightarrow (\psi \bullet_c \neg \mu) \wedge \mu \\
\psi \bullet_c \mu &\leftrightarrow \psi \vee (\psi \diamond_c \neg \mu).
\end{aligned}
$$

---

8. We note however that these approaches differ. Specifically, the PMA update operator satisfies all of the KM postulates, whereas our operator does not; see Section 4 for details.

### 3.3 Erasure

In Definition 8 we defined a dual to update, called *erasure*, directly in terms of *UL*. We can equally well define a function analogous to *UL*, call it *EL*, to directly define an erasure operator from first principles. We do this now, toward such a definition of erasure. Briefly, our motivation is: if we want to erase $\mu$ as a consequence of $\psi$, then semantically we want to add interpretations to the models of $\psi$. If $\mu$ corresponds to a single literal, then for each model of $\psi$ we would want to add an interpretation in which $l$ was replaced by $\bar{l}$. If $\mu$ corresponds to a conjunction, then $\mu$ can be erased by erasing either of the conjuncts; if $\mu$ corresponds to a disjunction, then to erase $\mu$ both disjuncts must be erased. By continuing in this fashion we obtain the following definition:

**Definition 9** *For interpretation $\omega$ and finite $\Gamma \subseteq L$, define $EL(\omega, \Gamma)$ as follows:*

1. *If $\Gamma \subseteq Lits$ then*

$$EL(\omega, \Gamma) = \begin{cases} \{(\omega \downarrow \Gamma) \cup \overline{\Gamma}\} & \text{if } \bigvee \Gamma \not\vdash \top \\ \emptyset & \text{otherwise} \end{cases}$$

2. *If $\Gamma = \{\alpha \wedge \beta\} \cup \Gamma'$ then $EL(\omega, \Gamma) = EL(\omega, \{\alpha\} \cup \Gamma') \cup EL(\omega, \{\beta\} \cup \Gamma')$*

3. *If $\Gamma = \{\alpha \vee \beta\} \cup \Gamma'$ then $EL(\omega, \Gamma) = EL(\omega, \{\alpha, \beta\} \cup \Gamma')$*

4. *If $\Gamma = \{\neg(\alpha \vee \beta)\} \cup \Gamma'$ then $EL(\omega, \Gamma) = EL(\omega, \{\neg\alpha\} \cup \Gamma') \cup EL(\omega, \{\neg\beta\} \cup \Gamma')$*

5. *If $\Gamma = \{\neg(\alpha \wedge \beta)\} \cup \Gamma'$ then $EL(\omega, \Gamma) = EL(\omega, \{\neg\alpha, \neg\beta\} \cup \Gamma')$*

6. *If $\Gamma = \{\neg\neg\alpha\} \cup \Gamma'$ then $EL(\omega, \Gamma) = EL(\omega, \{\alpha\} \cup \Gamma')$*

The following results are analogous to Theorems 2 and 3; note the occurrence of *cnf* in Theorem 8, in contrast to *dnf* in Theorem 3.

**Theorem 7** $EL(\omega, \Gamma) = EL(\omega, nnf(\bigwedge \Gamma))$.

**Theorem 8** $EL(\omega, \Gamma) = EL(\omega, cnf(\bigwedge \Gamma))$.

We can now directly define an erasure operator $\blacksquare'_c$ in terms of *EL*:

**Definition 10** $Mod(\psi \blacksquare'_c \mu) = Mod(\psi) \cup \{\omega' \mid \omega' \in EL(\omega, \{\mu\}), \omega \in Mod(\psi)\}$

Unsurprisingly, this notion of erasure and that given in Definition 8 are equivalent. We show this by first establishing the following result:

**Lemma 1** *For interpretation $\omega$ and $\Gamma \subseteq L$, we have $EL(\omega, \Gamma) = UL(\omega, \{\neg \bigwedge \Gamma\})$*

¿From this, it follows that our notions of erasure as given via the Harper Identity, and by direct definition via Definition 9 coincide:

**Theorem 9** $\psi \blacksquare_c \mu \leftrightarrow \psi \blacksquare'_c \mu$.

Hence we just use the symbol $\blacksquare_c$ for erasure. As a corollary, Theorem 9 also establishes the well-definedness of $\blacksquare'_c$.[9]

---

9. That is, since *UL* is well-defined (Theorem 1), so is $\blacksquare_c$ (Definition 8) and hence so is $\blacksquare'_c$ by the above equivalence.

### 3.4 Revision

In this section we consider extending the compositional approach to belief revision. To begin, it might be pointed out that there is nothing about the underlying motivation that makes $\diamond_c$ an update operator, and this point suggests that $\diamond_c$ might also be regarded as a revision operator, albeit with weak properties. However, regardless of intuitions, the recursive decomposition implicit in Definition 6 yields an operator with update-like properties, in that for sentence for update $\mu$, one effectively deals with the models of the disjuncts in $dnf(\mu)$. For revision, in contrast, the intuition is that one deals with models of $\mu$ that are (in some sense) closest to those of the knowledge base $\psi$. Hence, the operator $\diamond_c$ is not really appropriate as a revision operator.

This suggests a possibly-feasible approach to defining compositional revision: To define a revision $\psi * \mu$, one first uses the operator $\diamond_c$ to find a candidate set of models of $\mu$, and then employs some distance function to determine the subset of these models that are closest to models of $\psi$ as a whole. That is, for formulas $\psi$, $\mu$, an update of $\psi$ by $\mu$ is defined (in one fashion or another) with respect to all the models of $\psi$. For revision in contrast, a definition of the revision of $\psi$ by $\mu$ makes reference to only a *subset* of the models of $\psi$, those that are closest (in some sense) to the models of $\mu$. In this sense then, update is a logically weaker operator than revision. Thus a revision operator can be defined with respect to $\psi$ and $\mu$ by first applying some (compositional) update operator to get a candidate set of models of $\mu$. This set can then be "filtered", by removing those models that are not of minimal distance to the closest models of $\psi$. So depending on the notion of distance employed, one might expect to obtain different revision operators for a given compositional update operator.

There are two common notions of distance that are used for model-based belief change, one based on set containment and the other on cardinality. In the first case, for formulas $\alpha, \beta$, define

$$\Delta^{\min}(\alpha, \beta) \quad = \quad \min_{\subseteq}(\{M_1 \Delta M_2 \mid M_1 \in Mod(\alpha), M_2 \in Mod(\beta)\}),$$

where for sets $A$ and $B$, $A \Delta B$ is the symmetric difference of $A$ and $B$. Satoh's (1988) revision operator $\psi *_S \mu$ is defined as follows.

**Definition 11**

$$Mod(\psi *_S \mu) \quad = \quad \{w' \in Mod(\mu) \mid \exists w \in Mod(\psi), w \Delta w' \in \Delta^{\min}(\psi, \mu)\}.$$

For example, let $\psi = a \wedge b \wedge c$ and let $\mu = \neg a \vee (\neg b \wedge \neg c)$. Then $\psi *_S \mu = (\neg a \wedge b \wedge c) \vee (a \wedge \neg b \wedge \neg c)$.

We can define a corresponding compositional revision operator as follows:

**Definition 12**

$$Mod(\psi * \mu) \quad = \quad \{w' \mid w' \in UL(w, \{w\}), \ where \ w \in Mod(\psi), w \Delta w' \in \Delta^{\min}(\psi, \mu)\}.$$

However, it turns out that this revision operator in fact coincides the Satoh revision operator:

**Theorem 10** $\psi * \mu \ \leftrightarrow \ \psi *_S \mu$.

It follows as a straightforward corollary that if we use a distance metric based on the number of differing propositional symbols between two interpretations, we obtain the revision operator of (Dalal, 1988).[10] So in the obvious approaches to compositional revision, we do not obtain new revision operators; which is to say, the recursive decomposition in the definition of $UL$ does not serve to select among models of $\mu$ in any interesting sense with respect to revision.

However, these considerations do lead to one interesting result, and that is they point the way to algorithms that may more efficiently compute the Satoh or Dalal revision: To compute the Satoh revision for example, one can use Definition 6 to determine a relevant subset of models of $\mu$, and then use $\Delta^{\min}(\psi, \mu)$ to determine the closest subset of these models to the set of models of $\psi$. As we discuss in Section 5, this initial filtering of models of $\mu$ may be done efficiently in certain syntactically-restricted cases.

## 4. Analysis of Compositional Update and Erasure

To start, we consider which of the Katsuno-Mendelzon update postulates our operator satisfies. We do not consider the set of corresponding compositional erasure postulates, since the results are analogous to those of the update postulates, and so are of limited additional interest. After considering the update postulates, we further explore the update and erasure operators, including properties resulting from the restriction of the syntactic form of the formula for update, and a comparison to related approaches.

**Theorem 11** $\diamond_c$ *satisfies* **U1**, **U3**, **U5**, **U7**, **U8**.

For a counterexample to **U2**, consider the first example given above, illustrating the approaches of Winslett, where for $\psi \diamond_c \mu$ we have $\psi = (b \wedge \neg m) \vee (\neg b \wedge m)$ and $\mu = b \vee m$. In our approach, for the update (as given in Definition 7) the first disjunct of $\mu$ viz., $b$, yields interpretations $\{b, \neg m\}$ and $\{b, m\}$ and the update by the second disjunct, $m$, gives interpretations $\{b, m\}$ and $\{\neg b, m\}$. Hence $\psi \diamond_c (b \vee m)$ is characterized by the interpretations $\{b, m\}$, $\{b, \neg m\}$, and $\{\neg b, m\}$ and so we get $\psi \diamond_c (b \vee m) \leftrightarrow (b \vee m)$. **U2** would dictate that the result be $\psi$; however, the above example suggests that **U2** is problematic in the context of update. To borrow an example from other works (Herzig & Rifi, 1999; Brewka & Herzberg, 1993), suppose an agent believes $p$ (that a certain coin shows heads). Now the world changes because of a toss of this coin (where the agent does not see the result). Letting $\neg p$ be that the coin shows tails, we note that the agent should believe $(p \vee \neg p)$. Yet note that $p \vdash (p \vee \neg p)$; so **U2** would stipulate that $p \diamond (p \vee \neg p)$ should be $p$, contrary to what we want. The operator $\diamond_c$, on the other hand, includes an additional model. This appears to make sense, because by updating by $b \vee m$ we are really telling the knowledge base that the world has changed so that one of $b \wedge m$ or $b \wedge \neg m$ or $\neg b \wedge m$ is true. Thus, in this case the update operator behaves like the *Gricean* belief change operator of Delgrande, Nayak, and Pagnucco (2005), where the goal is to incorporate *all and only* the new information.

---

10. That is, for fixed formulas, any model of the Dalal revision is a model of the Satoh revision. Rephrasing Definitions 11 and 12 for cardinality-based distance gives a result analogous to Theorem 10 for Dalal revision. We omit the details.

We note that we can modify our $\diamond_c$ operator in a simple fashion to satisfy **U2** as follows:[11]

$$\psi \diamond_c' \mu = \begin{cases} \psi & \text{if } \psi \vdash \mu \\ \psi \diamond_c \mu & \text{otherwise} \end{cases}$$

But for our purposes, although **U2** is indeed now satisfied, this modification sheds no light on our original goal of investigating ramifications of developing a compositional update operator, and so we do not further pursue this modification.

We next consider a counterexample for **U4**. Although $((\neg a \wedge b) \vee b) \leftrightarrow b$, nonetheless $Mod(a \diamond_c ((\neg a \wedge b) \vee b)) = \{\{\neg a, b\}, \{a, b\}\}$ while $Mod(a \diamond_c b) = \{\{a, b\}\}$. So **U4** is not satisfied since in our compositional approach parts of a sentence may provide implicit results not explicit in the sentence. Consider $(\neg a \vee b) \wedge (\neg b \vee c)$ to further illustrate this point. Updating by this sentence is effected by updating by the individual components, viz., $(\neg a \vee b)$ and $(\neg b \vee c)$. However, implicit in these parts is the fact that $(\neg a \vee c)$ is also true, and the addition of this (implied) sentence would affect the result of the update. We consider this behaviour further below.

A counterexample for **U6** is given by the following. Let

$$\begin{aligned} \psi &= a \vee b \\ \mu_1 &= (a \vee \neg a) \\ \mu_2 &= \top \end{aligned}$$

We have that

$$Mod((a \vee b) \diamond_c (a \vee \neg a)) = Mod(\top)$$

But we also have:

$$Mod((a \vee b) \diamond_c \top) = Mod(a \vee b)$$

So we have a case where $\psi \diamond_c \mu_1 \vdash \mu_2$ and also $\psi \diamond_c \mu_2 \vdash \mu_1$. Thus the antecedent conditions of **U6** are satisfied, but not $\psi \diamond_c \mu_1 \leftrightarrow \psi \diamond_c \mu_2$.

While $\diamond_c$ does not satisfy **U4** (substitution of logical equivalents) in general, it does satisfy some weaker conditions. First, our update obviously satisfies substitution of logical equivalents in the first argument of $\diamond_c$. As well, in light of Theorems 2 and 3, if $\mu_1$ and $\mu_2$ share the same negation normal form or disjunctive normal form, then they may be substituted one for the other as a formula for update. We summarize these results as follows:

**Observation 1**

1. If $\psi_1 \leftrightarrow \psi_2$ then $(\psi_1 \diamond_c \mu) \leftrightarrow (\psi_2 \diamond_c \mu)$.

2. If $nnf(\mu_1) = nnf(\mu_2)$ then $(\psi \diamond_c \mu_1) \leftrightarrow (\psi \diamond_c \mu_2)$.

3. If $dnf(\mu_1) = dnf(\mu_2)$ then $(\psi \diamond_c \mu_1) \leftrightarrow (\psi \diamond_c \mu_2)$.

---

11. Borgida (1985) employed a similar definition with respect to a revision operator.

Despite failing to satisfy some postulates (which, it can be noted, overlap with the postulates that Herzig & Rifi, 1999, think are undesirable), $\diamond_c$ does exhibit a nice property, reflecting the compositional nature of our operator, but which operators appearing in the literature and satisfying the Katsuno and Mendelzon postulates fail to satisfy. The following version of the disjunction property holds.

**Theorem 12** $\psi \diamond_c (\mu_1 \vee \mu_2) \leftrightarrow (\psi \diamond_c \mu_1) \vee (\psi \diamond_c \mu_2)$

**Corollary 1** $(\psi \diamond_c \mu_1) \wedge (\psi \diamond_c \mu_2)$ *implies* $\psi \diamond_c (\mu_1 \vee \mu_2)$.

The corollary can be observed to be a strengthening of **U7**.

Our update operator satisfies those postulates deemed desirable by Herzig and Rifi (1999), with the exception of **U4**. As discussed above, **U4** is not satisfied due to the interaction of parts of a sentence. It would seem that if we could "compile out" the implicit information in a sentence then we would obtain the full substitution of equivalents, as expressed in **U4**. So, one way to satisfy **U4** is to redefine $\diamond_c$ so that we first get this information implicit in the interaction of the compositionally distinct parts of the update sentence. We do this by defining operators that consider the set of *prime implicants* of a sentence. We call this modified operator $\diamond_c^{pi}$. Let $PI(\mu)$ be the set of prime implicants of $\mu$.

**Definition 13** $\psi \diamond_c^{pi} \mu = \psi \diamond_c \bigvee PI(\mu)$

**Theorem 13** $\diamond_c^{pi}$ *satisfies* **U4**

Although $\diamond_c^{pi}$ satisfies **U4**, we now lose **U7**. A counter-example for **U7** is given by

$$\begin{aligned} \psi &= a \wedge b \wedge c \wedge d \\ \mu_1 &= (a \wedge d) \vee (\neg c \wedge d) \\ \mu_2 &= (\neg a \wedge d) \vee (\neg c \wedge d). \end{aligned}$$

We have that

$$\begin{aligned} Mod(\psi \diamond_c^{pi} \mu_1) &= \{\{a, b, c, d\}, \{a, b, \neg c, d\}\} \qquad \text{and} \\ Mod(\psi \diamond_c^{pi} \mu_2) &= \{\{\neg a, b, c, d\}, \{a, b, \neg c, d\}\} \end{aligned}$$

Hence $Mod(\psi \diamond_c^{pi} \mu_1) \cap Mod(\psi \diamond_c^{pi} \mu_2) = \{\{a, b, \neg c, d\}\}$. On the other hand

$$\begin{aligned} Mod(\psi \diamond_c^{pi} (\mu_1 \vee \mu_2)) &= Mod(\psi \diamond_c^{pi} ((a \wedge d) \vee (\neg c \wedge d) \vee (\neg a \wedge d) \vee (\neg c \wedge d))) \\ &= Mod(\psi \diamond_c^{pi} d) \\ &= \{\{a, b, c, d\}\}. \end{aligned}$$

Conversion to prime implicants in effect removes irrelevant or redundant syntactic information, as illustrated in the preceding example where $\mu_1 \vee \mu_2$ was in fact equivalent to the atom $d$. We can further pursue this line of inquiry by considering, for a formula for update $\mu$, a syntactic representation of the *proposition* expressed by $\mu$ over the language of $\mu$. For a given formula $\mu$, recall that $ModL(\mu)$ is the set of models of $\mu$, over the language of $\mu$. The formula $\bigvee\bigwedge ModL(\mu)$ then would be this formula expressed in disjunctive normal form; for example $ModL((a \vee b) \wedge c)$ would be expressed as $(a \wedge b \wedge c) \vee (a \wedge \neg b \wedge c) \vee (\neg a \wedge b \wedge c)$.

We define an update operator as follows:

**Definition 14** $Mod(\psi \diamond_c^{ss} \mu) = Mod(\psi \diamond_c (\bigvee\bigwedge ModL(\mu)))$.

We obtain that this update operator is in fact the same as that of Winslett's *standard semantics*:

**Theorem 14** $\psi \diamond_c^{ss} \mu \leftrightarrow \psi \diamond_{ss} \mu$.

We can pursue this direction one step further, and define an update operator where the update formula $\mu$ is characterized by its models expressed in dnf. That is we can define:

$$Mod(\psi \diamond_c^{triv} \mu) = Mod(\psi \diamond_c (\bigvee\bigwedge Mod(\mu))).$$

This is the same as Definition 14, except over models of $\mu$, rather than models of $\mu$ in the language of $\mu$. However it is easily shown that this is not an interesting operator, since it removes all old information and we have:

**Observation 2** $(\psi \diamond_c^{triv} \mu) \leftrightarrow \mu$.

We next proceed in a slightly different direction and compare our update operator with the *forget* operator. Recall that *forget*ting a set of atoms from a formula $\psi$ basically removes all information having to do with this set of atoms; in a sense forgetting is analogous to decreasing the language by this set of atoms.

To begin with, it can be noted that our update operator can have contraction-like properties similar to *forget*. For example, $(a \wedge b) \diamond_c (a \vee \neg a)$ is readily shown to be equivalent to $b$. So again, in a sense, this update can be read as updating by *precisely* $a \vee \neg a$, which in this case would indicate tautologous information concerning $a$. In fact, we have the following result (recall that we use $\odot$ to denote *forget*):

**Theorem 15** *Let $\psi \in L$ and let $\Gamma \subseteq \mathbf{P}$. Then*

$$\psi \odot \Gamma = \psi \diamond_c \left( \bigwedge_{p \in \Gamma} (p \vee \neg p). \right)$$

Hence forgetting a set of atoms is a special case of our update operator. (Given Theorem 6, *forget* is of course also expressible via erasure in an analogous fashion.) Last, we establish a result between the *forget* operator and Winslett's standard semantics. While in hindsight obvious, this result does not appear to have been previously noted.

**Theorem 16** *For formula $\psi$ and $\Gamma \subseteq \mathbf{P}$, let $\mu = \bigwedge_{l \in \Gamma} (l \vee \neg l)$. Then*

$$\psi \odot \Gamma = \psi \diamond_c^{ss} \mu.$$

In summary, it can be observed from the above discussion that we have obtained a hierarchy of operators, based on the extent to which information in $\mu$ is made explicit. For the most basic case, we have $\psi \diamond_c^{triv} \mu$, where the update formula $\mu$ is a syntactic representation of all models of the language; a trivial update operator results. The most basic interesting operator is given by $\psi \diamond_c^{ss} \mu$, which is the same as Winslett's standard semantics, followed by $\psi \diamond_c^{pi} \mu$ and $\psi \diamond_c \mu$. As well, by introducing tautologies, we also capture the notion of forgetting of atoms.

We have already noted that our update operator $\diamond_c$ is distinct from the Winslett PMA approach. To the best of our knowledge it is also distinct from all other specific approaches appearing in the literature, including those surveyed in (Herzig & Rifi, 1999).

## 5. Algorithms and Complexity

In this section we present a syntactic characterization as well as an algorithm for computing compositional update. We also analyze the complexity of this algorithm under a variety of assumptions. Specifically, we analyze the complexity of the algorithm when applied to any propositional sentences in general, to any sentences in disjunctive normal form, and to any sentences whose sizes are bounded by some specified constant.

We start with some background notions. Recall that $\psi[p/q]$ denotes the formula $\psi$ where all occurrences of atom $p$ are replaced by $q$. We write $\exists p.\psi$ to denote the formula $\psi[p/\top] \vee \psi[p/\bot]$. If $P = \{p_1, \cdots, p_n\}$ is a set of atoms then $\exists P.\psi$, called an *eliminant* of $P$ in $\psi$, stands for $\exists p_1.(\cdots \exists (p_n.\psi))$ (Brown, 1990). Intuitively, an eliminant of $P$ in $\psi$ can be viewed as a formula representing the same knowledge of $\psi$ that is not concerned with atoms in $P$. We have eliminated information about members of $P$ by replacing them by their two possible values, $\top$ and $\bot$, thus leaving only the other information in $\psi$.

It has been shown that Winslett's standard semantics can be syntactically captured based on the notion of eliminant (Doherty, Łukaszewicz, & Madalińska-Bugaj, 1998). Let $P = atom(\mu)$, then

$$\psi \diamond_{ss} \mu \leftrightarrow (\exists P.\psi) \wedge \mu \tag{5}$$

### 5.1 Syntactic Characterization and Algorithms

We are now ready to provide a syntactical characterization of compositional update. The idea is quite similar to that of (Doherty et al., 1998). However, our approach first converts the update formula to disjunctive normal form, then deals with each disjunct.

$$Update(\psi, \mu) = \bigvee \{(\exists P.\psi) \wedge t \mid t \in dnf(\mu), P = atom(t)\}$$

The following results establish the correspondence between the semantic definition and syntactic characterizations of compositional update.

**Lemma 2** *Suppose $t$ is a term (a conjunction of literals) and $P = atom(t)$. Then*

$$Mod((\exists P.\psi) \wedge t) = \{w' \mid w' \in UL(w, t), w \in Mod(\psi)\}$$

**Theorem 17** $Mod(\psi \diamond_c \mu) = Mod(Update(\psi, \mu)).$

**Corollary 2** $\psi \diamond_c \mu \leftrightarrow Update(\psi, \mu)$

All we need to compute compositional update, therefore, is the ability to compute eliminants. As proposed by Brown (1990), an eliminant $\exists P.\psi$ can be constructed as follows.

1. Convert $\psi$ to dnf $t_1 \vee \cdots \vee t_n$ (each $t_i$ is a conjunction of literals)

2. Replace each $t_i$ by $t_i \downarrow P$.

Now we are ready to provide the algorithms for compositional update. We will assume that $dnf(\mu)$ refers to the disjunctive normal form of $\mu$ represented in clause form, in which a formula is represented by sets of sets of literals. In this case, the members of $dnf(\mu)$ are implicitly disjoined, while a set of literals making up a member of $dnf(\mu)$ is implicitly conjoined.

In the following algorithms, let $\psi, \mu \in L$ and $P$ be a set of atoms:

**Algorithm** $Eliminant(P, \psi)$
1.    $\psi' \leftarrow \bot$
2.    **for each** term $t \in dnf(\psi)$
3.        $t' \leftarrow \top$
4.        **for each** literal $l \in t$
5.            **if** $l \notin P$ **and** $\bar{l} \notin P$
6.                $t' \leftarrow t' \wedge l$
7.        $\psi' \leftarrow \psi' \vee t'$
8.    **return** $\psi'$

**Algorithm** $Update(\psi, \mu)$
1.    $\psi' \leftarrow \bot$
2.    **for each** term $t \in dnf(\mu)$
3.        $P = atom(t)$
4.        $\psi' \leftarrow \psi' \vee (Eliminant(P, \psi) \wedge t)$
5.    **return** $\psi'$

Let's consider again Example 1 in which $\mu = b$ and $\psi = (b \wedge \neg m) \vee (\neg b \wedge m)$. We have that $Update(\psi, \mu) = \bot \vee (Eliminant(\{b\}, \psi) \wedge b)$. Since $Eliminant(\{b\}, \psi) = \bot \vee (\top \wedge \neg m) \vee (\top \wedge m)$, which is equivalent to $\top$, we obtain that $Update(\psi, \mu) \leftrightarrow b$. Thus, $Update(\psi, \mu) \leftrightarrow \psi \diamond_c \mu$, as we have already shown that $\psi \diamond_c \mu \leftrightarrow b$.

For Example 2, where $\psi = \neg b \wedge \neg m$ and $\mu = (b \vee m)$, we obtain $Update(\psi, \mu) = (b \wedge \neg m) \vee (\neg b \wedge m)$. Again, this result is same as what we obtain with $\diamond_c$.

## 5.2 Complexity

In the sequel, we analyze the space complexity of the update algorithm; that is, we are interested in how large the updated knowledge base could be. Unfortunately, when applied to arbitrary formulas, the algorithm $Update$ may cause exponential space blowup, as the disjunctive normal form of a formula could be exponentially large.

**Theorem 18** *The space complexity of $Update(\psi, \mu)$ is $\mathrm{O}(2^{(|\psi| + |\mu|)})$ for $\psi, \mu \in L$;*

However, we are able to show that such exponential space blowup is inevitable for any algorithm of compositional update. To this end, we need to introduce so-called *advice-taking* Turing machine (TM) and *non-uniform* complexity class, see (Johnson, 1990).

An *advice-taking* TM is a TM with an *advice oracle*, which can be considered as a function $a$ from positive integers to strings. On input $x$, the machine loads string $a(|x|)$ and then continues as usual based on two inputs $x$ and $a(|x|)$. Note that the oracle string $a(|x|)$ only depends on the size of the input $x$. We call an advice oracle $a$ polynomial iff $|a(n)| < p(n)$ for some fixed polynomial $p$ and all positive integers $n$. If X is a usual complexity class defined in terms of resource-bounded machines (e.g., P or NP) then X/poly is the class of the problem that can be decided on machines with the same resource bound augmented by polynomial advice oracles. Any class X/poly is also known as the non-uniform $X$; in particular, P/poly appears to be much more powerful than P. However, it has been shown very unlikely that NP $\subseteq$ P/poly, otherwise the polynomial hierarchy would collapse at $\Sigma_2^p$ (Karp & Lipton, 1980). This result is used to show that it is unlikely that there exists an algorithm for compositional update with a polynomial space bound.

**Theorem 19** *Assume there exist a polynomial $p$ and an algorithm* Update *of compositional update such that* $Update(\psi, \mu) \leftrightarrow \psi \diamond_c \mu$ *and* $|Update(\psi, \mu)| \leq p(|\psi| + |\mu|)$, *for any belief base $\psi$ and formula $\mu$. Then* $NP \subseteq P/poly$.

We can pursue the above result one step further, and show that algorithms for any sensible update operators will cause exponential blowup. Formally, we say an update operator $\diamond$ is *sensible* iff for any consistent set of literals $\Gamma$:

$$Mod(\psi \diamond \bigwedge \Gamma) = \{\omega' \mid \omega' \in (w \downarrow \Gamma) \cup \Gamma, \omega \in Mod(\psi)\}$$

Arguably, the above condition is very intuitive and natural (cf. discussions in Section 3). In fact, almost all update operators in the literature are sensible.

**Theorem 20** *If there exists a polynomially space bounded algorithm for any sensible update operator, then $NP \subseteq P/poly$.*

We remark that the above result also proves Winslett's conjecture stating that there does not exist a polynomially space bounded algorithm for her standard semantics (see Winslett, 1990).

The algorithm becomes tractably better when applied to formulas in disjunctive normal form, and to update formulas whose sizes are bounded.

**Theorem 21** *The space complexity of $Update(\psi, \mu)$ is:*

1. *$O(|\psi| \times |\mu|)$ for $\psi, \mu$ in dnf; and*

2. *$O(|\psi|)$ for $\psi$ in dnf and $|\mu| < k$ for some constant $k$.*

Arguably, in practice, the update formula $\mu$ (representing the changes of the world) will be relatively small. Therefore, it is relatively easy to convert $\mu$ to dnf, and it is also reasonable to assume the size of $\mu$ is bounded. As we usually do not restrict the size of the belief base $\psi$, converting $\psi$ to dnf could be computationally much more expensive. Fortunately, we only need to compile (off-line) the original belief base once into dnf, and the output of $Update$ algorithm is automatically the dnf of the updated belief base. This will considerably facilitate the further update of the belief base.

## 6. Conclusion

We have presented belief change operators for updating a knowledge base where the definition of these operators is compositional with respect to the sentence to be added. The intent is to provide operators with transparent definitions, based on the structure of the formula for belief change. As a result we lose some of the standard postulates for update, although we do satisfy a core group of the standard postulate set. We achieve full irrelevance of syntax if the sentence for update is replaced by the disjunction of its prime implicants.

The approach is interesting because first, it is founded on differing intuitions than other operators, in that it is based on a decomposition of the formula rather than on the models of the formula, and second, it allows a straightforward and (under reasonable assumptions) efficient implementation. While distinct from previous update operators that have appeared in the literature, we can capture Winslett's *standard semantics* approach to update in a restriction of our approach. In fact, the update operator, under different syntactic restrictions, may be regarded as constituting a family of update operators of which Winslett's *standard semantics* is the weakest interesting approach. When we turn from update to revision, we discover there is no new, interesting compositional revision operator; nevertheless, our results indicate that by first computing the compositional update, one can implement the Satoh or Dalal revision operator more efficiently, because we consider only a subset of the models of the formula of revision, and in certain cases this will have a significant speedup over a naive algorithm.

An open question concerns combining this approach with one that is designed to exploit the structure of the knowledge base (such as discussed in Parikh, 1999 and characterized in terms of PMA updates in Peppas, Chopra, & Foo, 2004). A second, technical question that is not fully explored concerns the behaviour of $\diamond_c$ as an erasure operator. For example, let $\psi = (a \vee b) \wedge (\neg a \vee \neg b)$. Then, we get that $\psi \diamond_c (a \vee b) \leftrightarrow a \vee b$. So, in updating the knowledge base with a formula already implied by the knowledge base, we have actually removed information. This, as discussed earlier, is quite reasonable if one considers that an update (in contrast to a revision) by $a \vee b$ asserts that the world has changed so that one of $\{a, b\}$, $\{\neg a, b\}$, $\{a, \neg b\}$ is now true. Finally, it would be of interest to apply the compositional approach to the merging of knowledge bases.

## Acknowledgments

An early precursor of this paper was presented at FLAIRS 2007 (Delgrande, Pelletier, & Suderman, 2007). We are grateful to that audience for comments; and we have also benefited from the perceptive comments of three JAIR referees. Delgrande and Pelletier also acknowledge the support of the Canadian NSERC granting agency.

## Appendix A. Proof of Theorems

**Proof 1.**

Observe that $UL$ is associative and commutative with respect to top-level conjunctions and top-level disjunctions. That is, for example

$$UL(\omega, \{\alpha \wedge (\beta \wedge \gamma)\} \cup \Gamma') = UL(\omega, \{(\alpha \wedge \beta) \wedge \gamma\} \cup \Gamma').$$

A similar observation can be made about negations of such top-level conjunctions and disjunctions; for example we have

$$UL(\omega, \{\neg(\alpha \wedge (\beta \wedge \gamma))\} \cup \Gamma') = UL(\omega, \{\neg((\alpha \wedge \beta) \wedge \gamma)\} \cup \Gamma').$$

We use such basic facts without comment in the sequel.

The above means in particular that for showing the order-independence of $UL$ with respect to its second argument, we just need consider the general case of $UL(\omega, \Gamma)$ where $\Gamma = \{\mu_1\} \cup \{\mu_2\}$, since we have that $UL(\omega, \{\alpha_1, \ldots, \alpha_n\}) = UL(\omega, \{\alpha_1, \bigwedge_{i=2}^{n} \alpha_i\})$.

Given this preamble, what we need to show is that for formulas $\mu_1$ and $\mu_2$, that $UL(\omega, \{\mu_1\} \cup \{\mu_2\})$ is independent of whether the initial recursion is in terms of $\mu_1$ or $\mu_2$.

The proof is on the *depth* of a formula.

**BASE:**

Assume that $depth(\mu_1) \leq 1$ and $depth(\mu_2) \leq 1$:

1. If $depth(\mu_1) = depth(\mu_2) = 0$ then $\mu_1$, $\mu_2$ are atoms and the result follows trivially.

2. If the only connective for $\mu_1$, $\mu_2$ is negation then $\mu_1$, $\mu_2$ are literals and again the result follows trivially.

3. If the connectives for $\mu_1$, $\mu_2$ are from $\{\neg, \wedge\}$, then $\mu_1$, $\mu_2$ reduce to sets of literals, and the previous case applies.

4. $\mu_1$ is $a_1 \vee a_2$, and $\mu_1$ is a literal, then only Step 3 of the definition applies, and our result obtains easily. The converse where $\mu_2$ is $a_1 \vee a_2$, and $\mu_2$ is a literal of course yields the same result.

5. If $\mu_1$ is $a_1 \wedge a_2$ and $\mu_2$ is $b_1 \vee b_2$, then we have

$$
\begin{aligned}
UL(\omega, \{a_1 \wedge a_2\} \cup \{\mu_2\}) &= UL(\omega, \{a_1, a_2\} \cup \{\mu_2\}) \\
&= UL(\omega, \{a_1, a_2\} \cup \{b_1 \vee b_2\}) \\
&= UL(\omega, \{b_1 \vee b_2\} \cup \{a_1, a_2\}) \\
&= UL(\omega, \{b_1\} \cup \{a_1, a_2\}) \cup UL(\omega, \{b_2\} \cup \{a_1, a_2\}) \\
&= UL(\omega, \{b_1\} \cup \{a_1 \wedge a_2\}) \cup UL(\omega, \{b_2\} \cup \{a_1 \wedge a_2\}) \\
&= UL(\omega, \{b_1 \vee b_2\} \cup \{a_1 \wedge a_2\}) \\
&= UL(\omega, \{b_1 \vee b_2\} \cup \{\mu_1\})
\end{aligned}
$$

6. If $\mu_1$ is $a_1 \vee a_2$ and $\mu_2$ is $b_1 \vee b_2$, then we have

$$
\begin{aligned}
UL(\omega, \{a_1 \vee a_2\} \cup \{\mu_2\}) &= UL(\omega, \{a_1\} \cup \{\mu_2\}) \cup UL(\omega, \{a_2\} \cup \{\mu_2\}) \\
&= UL(\omega, \{a_1\} \cup \{b_1 \vee b_2\}) \cup UL(\omega, \{a_2\} \cup \{b_1 \vee b_2\}) \\
&= UL(\omega, \{b_1 \vee b_2\} \cup \{a_1\}) \cup UL(\omega, \{b_1 \vee b_2\} \cup \{a_2\}) \\
&= (UL(\omega, \{b_1\} \cup \{a_1\}) \cup UL(\omega, \{b_2\} \cup \{a_1\})) \cup \\
&\qquad (UL(\omega, \{b_1\} \cup \{a_2\}) \cup UL(\omega, \{b_2\} \cup \{a_2\})) \\
&= UL(\omega, \{b_1, a_1\}) \cup UL(\omega, \{b_2, a_1\}) \cup \\
&\qquad UL(\omega, \{b_1, a_2\}) \cup UL(\omega, \{b_2, a_2\})
\end{aligned}
$$

Analogous manipulations show that $UL(\omega, \{b_1 \vee b_2\} \cup \{\mu_1\})$ yields the same result.

**STEP:**

For the induction hypothesis, assume that our result holds for $depth(\mu_1) \leq n$ and $depth(\mu_2) \leq n$. We show that the desired result obtains for $depth(\mu_1) \leq (n+1)$ and $depth(\mu_2) \leq (n+1)$.

**A:** Consider first where $depth(\mu_1) \leq n$ and $depth(\mu_2) = n+1$.

1. $\mu_2$ is of the form $\neg\neg\alpha$:

   $UL(\omega, \{\mu_1\} \cup \{\mu_2\})$ is the same as $UL(\omega, \{\mu_1\} \cup \{\alpha\})$, and our result follows by the induction hypothesis.

2. $\mu_2$ is $\alpha \wedge \beta$:

   $UL(\omega, \{\mu_1\} \cup \{\mu_2\}) = UL(\omega, \{\mu_1\} \cup \{\alpha \wedge \beta\}) = UL(\omega, \{\mu_1, \alpha, \beta\}) = UL(\omega, \{\alpha \wedge \beta\} \cup \{\mu_1\})$.

3. $\mu_2$ is $\alpha \vee \beta$:

   $UL(\omega, \{\mu_1\} \cup \{\mu_2\}) = UL(\omega, \{\mu_1\} \cup \{\alpha \vee \beta\}) = UL(\omega, \{\mu_1, \alpha\}) \cup UL(\omega, \{\mu_1, \beta\})$ while $UL(\omega, \{\mu_2\} \cup \{\mu_1\}) = UL(\omega, \{\alpha \vee \beta\} \cup \{\mu_1\}) = UL(\omega, \{\mu_1, \alpha\}) \cup UL(\omega, \{\mu_1, \beta\})$.

4. $\mu_2$ is $\neg(\alpha \wedge \beta)$ or $\mu_2$ is $\neg(\alpha \vee \beta)$:

   This is handled the same as $\alpha \vee \beta$ or $\alpha \wedge \beta$ respectively.

**B:** Consider next where $depth(\mu_1) = n+1$ and $depth(\mu_2) = n+1$.

1. $\mu_1$ or $\mu_2$ is $\neg\neg\alpha$:

   This is the same as $\mu_1$ or $\mu_2$ being $\alpha$, from which our result holds via the induction hypothesis.

2. $\mu_1$ or $\mu_2$ is $\alpha_1 \wedge \beta_1$:

   Assume without loss of generality that $\mu_1$ is $\alpha_1 \wedge \beta_1$.

   (a) $\mu_2$ is $\alpha_2 \wedge \beta_2$:

   $UL(\omega, \{\mu_1\} \cup \{\mu_2\}) = UL(\omega, \{\alpha_1 \wedge \beta_1\} \cup \{\alpha_2 \wedge \beta_2\}) = UL(\omega, \{\alpha_1, \beta_1, \alpha_2, \beta_2\}) = UL(\omega, \{\mu_2\} \cup \{\mu_1\})$.

(b) $\mu_2$ is $\alpha_2 \vee \beta_2$:

   This case is handled the same as in the base case, where $\alpha_1, \beta_1, \alpha_2, \beta_2$ are atoms.

(c) $\mu_2$ is $\neg(\alpha_2 \wedge \beta_2)$ or $\mu_2$ is $\neg(\alpha_2 \vee \beta_2)$:

   This is handled the same as $\alpha_2 \vee \beta_2$ or $\alpha_2 \wedge \beta_2$ respectively.

3. $\mu_1$ or $\mu_2$ is $\alpha \vee \beta$:

   The proof here is the same as in the base case, where $\alpha, \beta$ are atoms.

4. $\mu_1$ $(\mu_2)$ is $\neg(\alpha \wedge \beta)$ or $\mu_1$ $(\mu_2)$ is $\neg(\alpha \vee \beta)$:

   This is handled the same as $\alpha \vee \beta$ or $\alpha \wedge \beta$ respectively.

Since this covers all cases, our result follows by induction. □

**Proof 2.**

The proof follows straightforwardly from the observations that for arbitrary $\omega$, $\Gamma$, we have:

$$UL(\omega, \{\neg\neg\alpha\} \cup \Gamma) = UL(\omega, \{\alpha\} \cup \Gamma)$$
$$UL(\omega, \{\neg(\alpha \wedge \beta)\} \cup \Gamma) = UL(\omega, \{\neg\alpha \vee \neg\beta\} \cup \Gamma)$$
$$UL(\omega, \{\neg(\alpha \vee \beta)\} \cup \Gamma) = UL(\omega, \{\neg\alpha \wedge \neg\beta\} \cup \Gamma)$$

An induction argument establishes that the value of $UL$ doesn't change under conversion of elements of its second argument to negation normal form. □

**Proof 3.**

This result follows from the preceding, plus the fact that for arbitrary $\omega$, $\Gamma$ we have that: $UL(\omega, \{\alpha \wedge (\beta \vee \gamma)\} \cup \Gamma) = UL(\omega, \{(\alpha \wedge \beta) \vee (\alpha \wedge \gamma)\} \cup \Gamma)$; that is, $UL$ is invariant under distribution of conjunction over disjunction. □

**Proof 4.**

Proof is by induction on the maximum depth of a formula in $\Gamma$.

If the maximum depth is 0, then all members of $\Gamma$ are literals, and the result is immediate from Definition 6. Otherwise the induction hypothesis is that the result holds where the maximum depth of a formula in $\Gamma$ is $n$, and the step is easily shown by appeal to truth conditions in classical propositional logic. □

**Proof 5.**

Right-to-left: This is a corollary of Theorem 4.

Left-to-right:

For arbitrary $\Gamma$ we have by Theorem 3 that $UL(\omega, \Gamma) = UL(\omega, dnf(\Gamma))$.

Let $dnf(\Gamma) = \gamma_1 \vee \cdots \vee \gamma_n$ where each $\gamma_i$ is a conjunction of literals.

Via Definition 6 we have that $UL(\omega, dnf(\Gamma)) = UL(\omega, \{\gamma_1\}) \cup \cdots \cup UL(\omega, \{\gamma_n\})$. For each $\gamma_i$, if $\gamma_i$ contains a complementary pair of literals then $UL(\omega, \{\gamma_i\}) = \emptyset$; otherwise $UL(\omega, \{\gamma_i\}) \neq \emptyset$.

If we assume that $\Gamma \not\vdash \bot$, then there is some $\gamma_i$ with no complementary literals, consequently $UL(\omega, \{\gamma_i\}) \neq \emptyset$ and so $\emptyset \neq UL(\omega, dnf(\Gamma)) = UL(\omega, \Gamma)$.

Thus $\Gamma \not\vdash \bot$ implies $UL(\omega, \Gamma) \not\models \bot$, which was to be shown. □

**Lemma 3** $\psi \wedge \mu \vdash \psi \diamond_c \mu$.

**Proof of Lemma 3.** If $\psi \wedge \mu \vdash \bot$ then the result is immediate.

Consequently assume that $\psi \wedge \mu$ is satisfiable, and let $\omega \in Mod(\psi \wedge \mu)$. We show that $\omega \in Mod(\psi \diamond_c \mu)$. Given Definition 7, and since we already have $\omega \in Mod(\psi)$, we just need to show that $\omega \in UL(\omega, \{\mu\})$.

We have by assumption that $\omega \in Mod(\mu)$, whence (Theorem 3) $\omega \in Mod(dnf(\mu))$. Since $Mod(dnf(\mu)) = Mod(dnf(\mu_1)) \cup \cdots \cup Mod(dnf(\mu_n))$ for $dnf(\mu) = \mu_1 \vee \cdots \vee \mu_n$ we get $\omega \in Mod(dnf(\mu_i))$ for some disjunct $\mu_i$ of $dnf(\mu)$.

Since $\omega \in Mod(dnf(\mu_i))$ it follows from the definition of $UL$ that $\omega = UL(\omega, \{\mu_i\})$; hence $\omega \in Mod(dnf(\mu))$ and so $\omega \in UL(\omega, \{\mu\})$, which was to be shown.

□

**Proof 6.** The second part of the theorem follows immediately from Definitions 7 and 8.

For the first part: Since $\psi \wedge \mu \vdash \psi \diamond_c \mu$ (Lemma 3), we have

$$
\begin{aligned}
\psi \diamond_c \mu \quad &\leftrightarrow \quad (\psi \wedge \mu) \vee (\psi \diamond_c \mu) \\
&\leftrightarrow \quad (\psi \wedge \mu) \vee ((\psi \diamond_c \mu) \wedge \mu) \\
&\leftrightarrow \quad (\psi \vee (\psi \diamond_c \mu)) \wedge \mu \\
&\leftrightarrow \quad (\psi \blacksquare_c \neg\mu) \wedge \mu
\end{aligned}
$$

The last step applies the other part of the theorem, established above. □

**Proof 7.**

The proof is the same as that for Theorem 2 with minor modifications. □

**Proof 8.**

The proof is analogous to that of Theorem 3, and is omitted. □

**Proof of Lemma 1.**

The proof is straightforward, except setting up the induction is a bit fiddly. The induction is based on the maximum depth of a formula in $\Gamma$. For $\Gamma \subseteq L$, let $depth(\Gamma) = \max_{\mu \in \Gamma} depth(\mu)$. We then stipulate that $\Gamma$ precedes $\Gamma'$ in the ordering for the induction if $depth(\Gamma) < depth(\Gamma')$, or if $depth(\Gamma) = depth(\Gamma') = n$ and the number of formulas in $\Gamma$ of depth $n$ is less than that in $\Gamma'$.

**BASE:**

Let $\Gamma$ be a set of literals.

If $\bigvee \Gamma \vdash \top$ then $EL(\omega, \Gamma) = \emptyset = UL(\omega, \{\neg \wedge \Gamma\})$.

If $\bigvee \Gamma \nvdash \top$ then:

$$\begin{aligned}
EL(\omega, \Gamma) &= (\omega \downarrow \Gamma) \cup \overline{\Gamma} \\
&= (\omega \downarrow \overline{\Gamma}) \cup \overline{\Gamma} \\
&= UL(\omega, \overline{\Gamma}) \\
&= UL\left(\omega, \left\{\neg \bigwedge \Gamma\right\}\right)
\end{aligned}$$

**STEP:**

Assume that the result holds for the first $n$ sets of formulas in the ordering, and let $\mu$ be a formula of maximum depth in $\Gamma$. Let $\Gamma' = \Gamma \setminus \{\mu\}$.

If $\mu = \alpha \wedge \beta$, then

$$\begin{aligned}
EL(\omega, \Gamma) &= EL(\omega, \{\alpha \wedge \beta\} \cup \Gamma') \\
&= EL(\omega, \{\alpha\} \cup \Gamma') \cup EL(\omega, \{\beta\} \cup \Gamma') \\
&= EL(\omega, \{\wedge(\{\alpha\} \cup \Gamma')\}) \cup EL(\omega, \{\wedge(\{\beta\} \cup \Gamma')\}) \\
&= UL(\omega, \{\neg \wedge (\{\alpha\} \cup \Gamma')\}) \cup UL(\omega, \{\neg \wedge (\{\beta\} \cup \Gamma')\}) \\
&= UL(\omega, \{\neg\alpha \vee (\neg \wedge \Gamma')\}) \cup UL(\omega, \{\neg\beta \vee (\neg \wedge \Gamma')\}) \\
&= UL(\omega, \{\neg\alpha \vee (\neg \wedge \Gamma') \vee \neg\beta \vee (\neg \wedge \Gamma')\}) \\
&= UL(\omega, \{\neg\alpha \vee \neg\beta \vee (\neg \wedge \Gamma')\}) \\
&= UL(\omega, \{\neg(\alpha \wedge \beta \wedge (\wedge\Gamma'))\}) \\
&= UL(\omega, \{\neg(\wedge\Gamma)\}).
\end{aligned}$$

The change from *EL* to *UL* above is justified by the induction hypothesis; otherwise all steps are by definition of *UL* or *EL*, or simple manipulation.

If $\mu = \alpha \vee \beta$, then

$$\begin{aligned}
EL(\omega, \Gamma) &= EL(\omega, \{\alpha \vee \beta\} \cup \Gamma') \\
&= EL(\omega, \{\alpha, \beta\} \cup \Gamma') \\
&= UL(\omega, \{\neg \wedge (\{\alpha, \beta\} \cup \Gamma')\}) \\
&= UL(\omega, \{\neg(\wedge\Gamma)\}).
\end{aligned}$$

Again, the change from *EL* to *UL* above is justified by the induction hypothesis.

Other cases are handled analogously; their proofs are omitted.

Hence our result follows by induction. □

**Proof 9.**

$$\begin{aligned}
Mod(\psi \bullet_c \mu) &= Mod(\psi) \cup \{\omega' \mid \omega' \in UL(\omega, \{\neg\mu\}), \omega \in Mod(\psi)\} \\
&= Mod(\psi) \cup \{\omega' \mid \omega' \in EL(\omega, \{\mu\}), \omega \in Mod(\psi)\} \\
&= Mod(\psi \bullet'_c \mu)
\end{aligned}$$

The first and last steps above are justified by Definitions 8 and 10 respectively; the middle step follows from Lemma 1. □

**Proof 10.** It follows immediately from Definitions 11 and 12 that $Mod(\psi * \mu) \subseteq Mod(\psi *_S \mu)$.

To show the converse, we let $w' \in Mod(\psi *_S \mu)$ and show that $w' \in Mod(\psi * \mu)$.

Given that the Satoh revision operator satisfies irrelevance of syntax (**R4**), we can assume without loss of generality that $\mu$ is in dnf; i.e. $\mu = \mu_1 \vee \cdots \vee \mu_n$ where each $\mu_i$ is a conjunction of literals.

We have by assumption that $w' \in Mod(\psi *_S \mu)$; hence $\exists w \in Mod(\psi)$ such that $w \Delta w' \in \Delta^{\min}(\psi, \mu)$.

Since $w' \in Mod(\psi *_S \mu)$, we have that $w' \in Mod(\mu) = Mod(\mu_1 \vee \cdots \vee \mu_n)$. Thus there is a clause from $\mu$, $\mu_i$, such that $w' \in Mod(\mu_i)$. Assume without loss of generality that $\mu_i$ is subset-minimal among the sets of literals making up the disjuncts of $\mu$.

If we can show $w' \in UL(w, \{\mu_i\})$ then we will have shown that $w'$ satisfies the conditions to be a member of $Mod(\psi * \mu)$. We show this as follows. Let $U_i$ be the set of literals in $\mu_i$.

We have that for $l \notin U_i$ that $l \in w$ iff $l \in w'$ (since otherwise this would contradict $w \Delta w' \in \Delta^{\min}(\psi, \mu)$).

It follows that $(w \downarrow U_i) \cup U_i = w'$. But this means that $w' \in UL(w, U_i)$, or $w' \in UL(w, \{u_i\})$ and so $w' \in UL(w, \{\mu\})$.

Hence $w' \in Mod(\psi * \mu)$, which was to be shown. $\qquad \square$

**Proof 11.**

**U1:** By Theorem 4, $UL(\omega, \{\mu\}) \models \mu$ for every $\omega \in Mod(\psi)$, whence $Mod(\psi \diamond_c \mu) \subseteq Mod(\mu)$ or $\psi \diamond_c \mu \vdash \mu$.

**U3:** By assumption $\psi \nvdash \bot$, and so $Mod(\psi) \neq \emptyset$. Our result then follows immediately from Theorem 5 and Definition 7.

**U5:** If $Mod(\psi \diamond_c \mu) \cap Mod(\phi) = \emptyset$ then our result follows vacuously.

Otherwise, let $\omega \in Mod(\psi \diamond_c \mu) \cap Mod(\phi)$.

Since $\omega \in Mod(\psi \diamond_c \mu)$ then, by Theorem 3, there exists $\omega' \in Mod(\psi)$ and $\Gamma \subseteq Lits$ such that $\bigwedge \Gamma$ is a disjunct of $dnf(\mu)$ where $\omega = (\omega' \downarrow \Gamma) \cup \Gamma$.

Because $\omega \in Mod(\phi)$ there is a set of literals $\Gamma'$ such that $\bigwedge \Gamma'$ is a disjunct of $dnf(\phi)$ such that $\Gamma' \subseteq \omega$.

By definition, $\bigwedge \Gamma \cup \Gamma'$ is a clause in $dnf(\mu \wedge \phi)$. We note that $\omega = (\omega' \downarrow (\Gamma \cup \Gamma')) \cup (\Gamma \cup \Gamma')$ so by Definition 7 we have $\omega \in Mod(\psi \diamond_c (\mu \wedge \phi))$.

**U7:**

$$
\begin{aligned}
& Mod(\psi \diamond_c \mu_1) \cap Mod(\psi \diamond_c \mu_2) \\
\subseteq \; & Mod(\psi \diamond_c \mu_1) \cup Mod(\psi \diamond_c \mu_2) \\
\subseteq \; & Mod(\psi \diamond_c (\mu_1 \vee \mu_2))
\end{aligned}
$$

The last step follows from Theorem 3, using the fact that $dnf(\alpha \vee \beta) = dnf(\alpha) \vee dnf(\beta)$.

Hence $(\psi \diamond_c \mu_1) \wedge (\psi \diamond_c \mu_2)$ implies $\psi \diamond_c (\mu_1 \vee \mu_2)$.

**U8:**

$$
\begin{aligned}
Mod((\psi_1 \vee \psi_2) \diamond_c \mu) &= \{\omega' \mid \omega' \in UL(\omega, \{\mu\}), \omega \in Mod(\psi_1 \vee \psi_2)\} \\
&= \{\omega' \mid \omega' \in UL(\omega, \{\mu\}), \omega \in Mod(\psi_1) \cup Mod(\psi_2)\} \\
&= \{\omega' \mid \omega' \in UL(\omega, \{\mu\}), \omega \in Mod(\psi_1)\} \\
&\qquad \cup \{\omega' \in UL(\omega, \{\mu\}), \omega \in Mod(\psi_2)\} \\
&= Mod(\psi_1 \diamond_c \mu) \cup Mod(\psi_2 \diamond_c \mu).
\end{aligned}
$$

From $Mod((\psi_1 \vee \psi_2) \diamond_c \mu) = Mod(\psi_1 \diamond_c \mu) \cup Mod(\psi_2 \diamond_c \mu)$ it follows that $((\psi_1 \vee \psi_2) \diamond_c \mu) \leftrightarrow (\psi_1 \diamond_c \mu) \vee (\psi_2 \diamond_c \mu)$.

$\square$

**Proof 12.**

$$
\begin{aligned}
Mod(\psi \diamond_c (\mu_1 \vee \mu_2)) &= \{\omega' \mid \omega' \in UL(\omega, \{\mu_1 \vee \mu_2\}), \omega \in Mod(\psi)\} \\
&= \{\omega' \mid \omega' \in UL(\omega, \{\mu_1\}), \omega \in Mod(\psi)\} \cup \\
&\qquad \{\omega' \mid \omega' \in UL(\omega, \{\mu_2\}), \omega \in Mod(\psi)\} \\
&= Mod(\psi \diamond_c \mu_1) \cup Mod(\psi \diamond_c \mu_2) \\
&= Mod((\psi \diamond_c \mu_1) \vee (\psi \diamond_c \mu_2))
\end{aligned}
$$

$\square$

**Proof 13.**

We need to show that if $\psi_1 \leftrightarrow \psi_2$ and $\mu_1 \leftrightarrow \mu_2$ then we have that $(\psi_1 \diamond_c \mu_1) \leftrightarrow (\psi_2 \diamond_c \mu_1)$. Since $\mu_1 \leftrightarrow \mu_2$ by assumption, we have that $PI(\mu_1) = PI(\mu_2)$.[12] Since $(\psi_1 \diamond_c \bigvee PI(\mu_1)) \leftrightarrow (\psi_2 \diamond_c \bigvee PI(\mu_1))$ and $(\psi_2 \diamond_c \bigvee PI(\mu_1)) \leftrightarrow (\psi_2 \diamond_c \bigvee PI(\mu_2))$, we have $(\psi_1 \diamond_c \bigvee PI(\mu_1)) \leftrightarrow (\psi_2 \diamond_c \bigvee PI(\mu_2))$, whence $(\psi \diamond_c^{pi} \mu_1) \leftrightarrow (\psi \diamond_c^{pi} \mu_2)$.

$\square$

**Proof 14.**

We have that $Mod(\psi \diamond_c^{ss} \mu) = Mod(\psi \diamond_c (\bigvee(\bigwedge ModL(\mu))))$. Using Definition 7, the right hand side is equal to $\{\omega' \mid \omega' \in UL(\omega, \{\bigvee(\bigwedge ModL(\mu))\}), \omega \in Mod(\psi)\}$. Hence, each $\omega \in Mod(\psi)$ is replaced by a set of interpretations $\Omega$ where $\Omega = \{(\omega \downarrow \omega'') \cup \omega'' \mid \omega'' \in ModL(\mu)\}$. Which is to say, $\omega \in Mod(\psi)$ is replaced by a set of interpretations $\Omega$ where $\omega'' \in \Omega$ just if $\omega$ and $\omega''$ differ only over the language of $\mu$. But this is just the definition for $Mod(\psi \diamond_c^{ss} \mu)$.

$\square$

**Proof 15.**

Let $\Gamma = \{p_1, \ldots, p_n\}$.

$$
\begin{aligned}
\psi \odot \Gamma &= \psi \odot \{p_1 \wedge \cdots \wedge p_n\} \\
&= [(\psi \odot \{p_1 \wedge \cdots \wedge p_n\}) \diamond_c p_n] \vee [(\psi \odot \{p_1 \wedge \cdots \wedge p_n\}) \diamond_c \neg p_n]
\end{aligned}
$$

---

12. Equality isn't quite right here. Rather we have equality modulo associativity and commutativity, which is all that we need for our result.

The second step above is just Definition 3 for forget expressed in terms of update. Definition 3 can be successively reapplied to eventually terminate with a disjunction with $2^n$ disjuncts, where each disjunct is a sequence of $n$ updates of literals from $\Gamma$. Moreover, every maximum consistent set of literals from $\Gamma$ appears in some disjunct.

We have the easy result, that we state without proof, that for disjoint sets of literals $\Gamma_1$, $\Gamma_2$, that $(\psi \diamond_c (\wedge\Gamma_1)) \diamond_c (\wedge\Gamma_2) = (\psi \diamond_c \wedge(\Gamma_1 \cup \Gamma_2))$.

Hence we get finally that

$$\psi \odot \Gamma \;=\; \bigwedge_{\Lambda \subseteq \Gamma} \psi \diamond_c ((\wedge\Lambda) \;\wedge\; \neg \wedge (\Gamma \setminus \Lambda))$$

$$=\; \psi \diamond_c \left( \bigwedge_{p \in \Gamma} (p \vee \neg p). \right)$$

□

**Proof 16.**

¿From Theorem 15 we have that

$$\psi \odot \Gamma \;=\; \psi \diamond_c \left( \bigwedge_{a \in \Gamma} (a \vee \neg a). \right) \;=\; \psi \diamond_c \mu.$$

¿From Observation 1 we get that $\psi \diamond_c \mu = \psi \diamond_c dnf(\mu)$. An easy argument shows that $dnf(\mu) = \bigvee \bigwedge ModL(\mu)$, and so Definition 14 yields $\psi \diamond_c dnf(\mu) = \psi \diamond_c^{ss} \mu$. Theorem 14 is $\psi \diamond_c^{ss} \mu = \psi \diamond_{ss} \mu$. Putting this all together we get $\psi \odot \wedge(\Gamma) = \psi \diamond_c^{ss} \mu$. □

**Proof of Lemma 2.**

Equation (5) implies that $Mod((\exists P.\psi) \wedge t) = Mod(\psi \diamond_{ss} t)$. According to Theorem 14 and Definition 14, we then have $Mod(\psi \diamond_{ss} t) = Mod(\psi \diamond_c (\bigvee(\bigwedge ModL(t)))) = Mod(\psi \diamond_c t)$. From Definition 7, it follows that $Mod((\exists P.\psi) \wedge t) = \{w' \mid w' \in UL(w,t), w \in Mod(\psi)\}$. □

**Proof 17.**

According to Definition 7, $Mod(\psi \diamond_c \mu) = \{w' \mid w' \in UL(w, \{\mu\}), w \in Mod(\psi)\}$. By Theorem 3, we have $\{w' \mid w' \in UL(w, \{\mu\}), w \in Mod(\psi)\} = \{w' \mid w' \in UL(w, \{dnf(\mu)\}), w \in Mod(\psi)\}$. From Definition 6, it follows that $Mod(\psi \diamond_c \mu) = \{w' \mid w' \in UL(w,t), t \in dnf(\mu), w \in Mod(\psi)\}$. According to Lemma 2, thus $Mod(\psi \diamond_c \mu) = Mod(Update(\psi, \mu))$. □

**Proof 18.**

The size of $dnf(\psi)$ is $O(2^{|\psi|})$. Hence, the size of $Eliminant(P, \psi)$ is also $O(2^{|\psi|})$. Similarly, the size of $dnf(\mu)$ is $O(2^{|\mu|})$. Therefore, $|Update(\psi, \mu)| = O(2^{|\mu|} \times 2^{|\psi|}) = O(2^{|\psi|+|\mu|})$. □

**Proof 19.**

This proof is inspired by the ideas in (Cadoli, Donini, Liberatore, & Schaerf, 1995), where it was shown that many revision operators cause exponential blowup. We show that if there exists a polynomially space bounded algorithm of compositional update, then 3SAT is in P/poly.[13] The proof consists of two steps.

**STEP 1:**

For any integer $n$, we first construct a belief base $\psi_n$ and a formula $\mu_n$, whose sizes are polynomial wrt. $n$. Let $X = \{x_1, \cdots, x_n\}$ and $Y = \{y_1, \cdots, y_n\}$ be two disjoint set of atoms and let $C$ be a set of new atoms for each 3-literal clause over $X$, i.e., $C = \{c_i \mid \gamma_i$ is a 3-literal clause of $X\}$. We obtain $\psi_n$ and $\mu_n$ as follows:

$$\psi_n = \{\gamma_i \vee \neg c_i \mid \gamma_i \text{ is a 3-literal clause of } X\}$$
$$\mu_n = \bigwedge_{i=1}^{n}(\neg x_1 \wedge \neg y_i)$$

It is easy to see that $|\psi_n| \in O(n^3)$ and $|\mu_n| \in O(n)$.

Then we show that for any 3CNF $\beta$ of size $n$, there exists an interpretation $\omega_\beta$ (on atoms $X \cup Y \cup C$) such that $\omega_\beta \models \psi_n \diamond_c \mu_n$ iff $\beta$ is satisfiable. We assume, without loss of generality, that $atom(\beta) \subseteq X$; or otherwise, we can always substitute atoms of $\beta$ respectively by elements of $X$ to obtain a new sentence $\beta_X$ such that $\beta$ is satisfiable iff $\beta_X$ is satisfiable. Then $w_\beta$ can be obtained as follows:

$$w_\beta = \{c_i \in C \mid \gamma_i \text{ is a clause of } \beta\} \cup \{\neg c_i \in C \mid \gamma_i \text{ is not a clause of } \beta\} \cup \overline{X} \cup \overline{Y}$$

We now show that $\beta$ is satisfiable iff $\omega_\beta \models \psi_n \diamond_c \mu_n$.

$\Rightarrow$ Assume $\beta$ is satisfiable. Let $\omega$ be a model of $\beta$. We construct another interpretation $\omega' = UL(\omega, \{\neg c_i \in C \mid \gamma_i$ is not a clause of $\beta\})$. It is easy to see that $\omega' \models \psi_n$ and $\omega_\beta = UL(\omega', \{\neg x_i, \neg y_i \mid 1 \le i \le n\})$. It follows that $\omega_\beta \models \psi_n \diamond_c \mu_n$.

$\Leftarrow$ Assume $\omega_\beta \models \psi_n \diamond_c \mu_n$. Then there exists an interpretation $\omega$ such that $\omega \models \psi_n$ and $\omega_\beta = UL(\omega, \{\neg x_i, \neg y_i \mid 1 \le i \le n\})$. We claim that $\omega \models \beta$. Assume $\omega \not\models \beta$. Then there exists a 3-literal clause $\gamma_i$ of $\beta$ such that $\omega \not\models \gamma_i$. From $\omega_\beta = UL(\omega, \{\neg x_i, \neg y_i \mid 1 \le i \le n\})$ and $c_i \in \omega_\beta$, if follows that $c_i \in \omega$. This implies $\omega \not\models \gamma_i \vee \neg c_i$, which contradicts $\omega \models \psi_n$. Thus, $\beta$ is indeed satisfiable.

**STEP 2:**

Suppose $Update$ is a polynomial space bounded algorithm of compositional update. Then 3SAT can be solved by an advice taking TM as follows: Given an arbitrary 3CNF $\beta$ of size $n$, the machine first loads the advice string $Update(\psi_n, \mu_n)$ and computes (in polynomial time) $\omega_\beta$ ; then it verifies $\omega_\beta \models Update(\psi_n, \mu_n)$. Since $|\psi_n| \in O(n^3)$, $|\mu_n| \in O(n)$, and $|Update(\psi_n, \mu_n)| \le p(|\psi_n| + |\mu_n|)$, we can do the verification in polynomial time. Since $Update(\psi_n, \mu_n) \leftrightarrow \psi_n \diamond_c \mu_n$, we have $\omega_\beta \models \psi_n \diamond_c \mu_n$ iff $\omega_\beta \models Update(\psi_n, \mu_n)$. Therefore, $\beta$ is satisfiable iff $\omega_\beta \models Update(\psi_n, \mu_n)$. This shows that 3SAT $\in$ P/poly. As 3SAT is NP-complete, we have NP $\subseteq$ P/poly.

$\square$

**Proof 20.**

---

13. A 3-literal clause is clause consists of precisely 3 literals and a 3CNF is a conjunction of 3-literal clauses. 3SAT is the satisfiability problem for 3CNFs, which has been shown NP-complete.

This proof is exactly same as that of Theorem 19, as the update formula $\mu_n$ used there is a consistent conjunction of literals. □

**Proof 21.** Since $\psi, \mu$ are in dnf, $|\psi| = |dnf(\psi)|$ and $|\mu| = dnf(\mu)$. Thus $|Eliminant(P, \psi)| = O|\psi|$. Therefore $|Update(\psi, \mu)| = O(|\psi| \times |\mu|)$.

In case $|\mu| < k$, we have $|Update(\psi, \mu)| = O(|\psi| \times k) = O(\psi)$.

□